\documentclass[10pt,letterpaper]{article}
\usepackage[top=0.85in,left=2.75in,footskip=0.75in]{geometry}

\usepackage{amsmath,amssymb}

\usepackage{changepage}

\usepackage[utf8x]{inputenc}

\usepackage{textcomp,marvosym}

\usepackage{cite}

\usepackage{nameref,hyperref}

\usepackage[right]{lineno}

\usepackage{microtype}
\DisableLigatures[f]{encoding = *, family = * }

\usepackage[table]{xcolor}

\usepackage{array}

\newcolumntype{+}{!{\vrule width 2pt}}

\newlength\savedwidth

\raggedright
\usepackage[aboveskip=1pt,labelfont=bf,labelsep=period,justification=raggedright,singlelinecheck=off]{caption}

\makeatletter
\renewcommand{\@biblabel}[1]{\quad#1.}
\makeatother

\usepackage{lastpage,fancyhdr,graphicx}
\usepackage{epstopdf}
\fancyheadoffset[L]{2.25in}
\fancyfootoffset[L]{2.25in}
\usepackage{booktabs}
\usepackage{float}
\usepackage{gensymb}
\usepackage{multirow}
\usepackage{url}
\begin{document}
\vspace*{0.2in}

% Title must be 250 characters or less.
\begin{flushleft}
  {\Large
    \textbf\newline{The Effects of Spectral Dimensionality Reduction on Hyperspectral Pixel Classification: A Case Study}
    % Please use "sentence case" for title and headings (capitalize only the first word in a title (or heading), the first word in a subtitle (or subheading), and any proper nouns).
  }
  \newline
  % Insert author names, affiliations and corresponding author email (do not include titles, positions, or degrees).
  \\
  Kiran Mantripragada\textsuperscript{1*},
  Phuong D. Dao\textsuperscript{2},
  Yuhong He\textsuperscript{2},
  Faisal Z. Qureshi\textsuperscript{1}
  \\
  \bigskip
  \textbf{1} Faculty of Science\\
  University of Ontario Institute of Technology\\
  2000 Simcoe Street North, Oshawa ON L1G OC5, Canada
  \\
  \textbf{2} Department of Geography, Geomatics and Environment\\
  University of Toronto\\
  3359 Mississauga Road, Mississauga ON L5L 1C6, Canada
  \\
  \bigskip

  % Insert additional author notes using the symbols described below. Insert symbol callouts after author names as necessary.
  %
  % Remove or comment out the author notes below if they aren't used.
  %
  % Primary Equal Contribution Note
  % \Yinyang These authors contributed equally to this work.

  % Additional Equal Contribution Note
  % Also use this double-dagger symbol for special authorship notes, such as senior authorship.
  % \ddag These authors also contributed equally to this work.

  % Current address notes
  % \textcurrency Current Address: Dept/Program/Center, Institution Name, City, State, Country % change symbol to "\textcurrency a" if more than one current address note
  % \textcurrency b Insert second current address
  % \textcurrency c Insert third current address

  % Deceased author note
  % \dag Deceased

  % Group/Consortium Author Note
  % \textpilcrow Membership list can be found in the Acknowledgments section.

  % Use the asterisk to denote corresponding authorship and provide email address in note below.
  * kiran.mantripragada@ontariotechu.net

\end{flushleft}
% Please keep the abstract below 300 words

\begin{abstract}
  This paper presents a systematic study of the effects of hyperspectral pixel dimensionality reduction on the pixel classification task.  We use five dimensionality reduction methods---PCA, KPCA, ICA, AE, and DAE---to compress 301-dimensional hyperspectral pixels.  Compressed pixels are subsequently used to perform pixel classifications.  Pixel classification accuracies together with compression method, compression rates, and reconstruction errors provide a new lens to study the suitability of a compression method for the task of pixel classification. We use three high-resolution hyperspectral image datasets, representing three common landscape types (i.e. urban, transitional suburban, and forests) collected by the Remote Sensing and Spatial Ecosystem Modeling laboratory of the University of Toronto. We found that PCA, KPCA, and ICA post greater signal reconstruction capability; however, when compression rates are more than 90\% these methods show lower classification scores. AE and DAE methods post better classification accuracy at 95\% compression rate, however their performance drops as compression rate approaches 97\%.  Our results suggest that both the compression method and the compression rate are important considerations when designing a hyperspectral pixel classification pipeline.
\end{abstract}

\section{Introduction}

There is a growing demand for improved hyperspectral image analysis in part
due to increasing availability of images with high spatial and spectral
resolutions~\cite{ghamisi:2017,lu:2020}. Hyperspectral images capture information
from the ultraviolet, visible, and infrared regions of the electromagnetic waves and
record the spectral signature of the observed
objects. The richness of information in the spectra is helpful in a variety
of tasks such as object detection, segmentation, and classification.
Consequently, hyperspectral images have found wide-spread use in a
number of application domains, such as mining, environmental monitoring, military, etc.
Unlike ordinary images, in which each pixel consists of $3$ channels (red, green, and blue),
a pixel in a hyperspectral image can consist of upwards of $300$ spectral data values.
This suggests that hyperspectral images have much higher
requirements in terms of storage space and computational processing.
Therefore, the search for better methods for hyperspectral image storage, processing,
and analysis continues unabated.

Spectral information stored at each pixel is often redundant~\cite{guo:2006},
therefore it is often not necessary to process
all spectral bands when performing hyperspectral image segmentation or
classification.  This is especially true when hyperspectral images
represent regions that contain a particular set of materials, often
referred to as {\it endmembers}~\cite{du:2003}.  A first step towards
developing computationally efficient techniques for hyperspectral
image processing is to reduce the inherent redundancy in hyperspectral images,
thereby reducing the amount of data that needs to be processed in the
subsequent steps. Within this context, feature extraction, band selection,
and compression have grown into active research areas.  A number of researchers
have explored dimensionality reduction techniques, such as Principal Component
Analysis (PCA), Kernel PCA (KPCA), and Independent Component Analysis (ICA),
to compress hyperspectral image pixels with a view to reducing the redundancy
inherent in these images~\cite{cao:2003,du:2003,rasti:2018}.
Autoencoder models have also been used to construct low-dimensional
features that are subsequently used for image analysis
tasks~\cite{zhou:2019,belwalkar:2018,ball:2018}.

The compression techniques used in hyperspectral image analysis aim to
find the lower-dimensional encoding of the spectral signal by
minimizing the reconstruction loss. This is a logical choice since it
ensures that the low-dimensional encoding retains the important
information needed to reconstruct the original spectral signal with
minimal distortion or loss.  We instead argue that a better approach to selecting the best compression method is to study the performance of the penultimate task---in our case, pixel classification---on the compressed signal.  Specifically, we seek a compression method that encodes the spectral signal in a low-dimensional space such that the low-dimensional encoding both
minimizes the reconstruction loss and maximizes the pixel classification performance.
This idea is motivated by lossy image compression approaches, e.g.,
the Joint Photographic Experts Group (JPEG) standard, that balance
perceptual loss against compression rates.  Note that this work is concerned with pixel level compression.  Image level compression should also exploit spatial information encoded within neighbouring pixels, and this work does not consider spatial information.

We setup the problem as follows.  First, we assert that the performance
of the final task that we want to carry out is a better proxy for evaluating
the performance of an image compression algorithm.  A central objective of
many hyperspectral image analysis methods is to achieve better pixel-level
classification; therefore, we decided to use pixel classification to study the
performance of five widely used compression techniques for hyperspectral pixels: three
dimensionality reduction methods---PCA, KPCA, and ICA--and two deep learning based approaches---Autoencoder (AE) and denoising autoencoder (DAE)---construct low-dimensional encodings of the input pixel at various compression rates ranging from $1\%$ to $99\%$.  These encodings are subsequently used for pixel label classification.  To the best of our
knowledge, this is the first systematic study that captures the interplay between compression methods and rates and the task of hyperspectral pixel classification.  We use three new hyperspectal images, each representing a common landscape type
(i.e. urban, transitional suburban, and forests) collected by the Remote Sensing
and Spatial Ecosystem Modeling laboratory of the University of Toronto,
to carry out the experiments (see Section~\ref{sec:datasets}).
Those who are interested in computationally efficient hyperspectral pixel analysis will find our findings useful.

The paper makes the following contributions: it presents a first-of-its-kind study of the effects of hyperspectral pixel compression on pixel classification.  The paper studies classification performance when hyperspectral pixels are compressed using one of five compression methods using various rates of compression.  The results suggest that AE and DAE methods create pixel encodings that achieve best classification scores for compression rates around $95\%$.  Additionally, we find that widely-used denoising filters are not needed when using AE or DAE methods for pixel compression.   

The rest of this paper is organized as follows.  The next section briefly
summarizes prior work.  Section~\ref{sec:datasets} describes the three
datasets that we used to carry out the experiments.  Compression and
classification methods are presented in Section~\ref{sec:methodology},
followed by results in Section ~\ref{sec:results} and conclusions in
Section~\ref{sec:conclusion}.

\section{Background}
\label{sec:background}

Data compression and pixel level classification are important topics in hyperspectral
image analysis.\footnote{Pixel level classification is oftened called
    {\it semantic segmentation} in the wider computer vision literature.}  Hyperspectral
images store two orders of magnitude more information than an ordinary RGB image,
and it is desirable to compress these images to reduce storage requirements, improve
processing speeds, and lower computational requirements.  In addition it is sometimes
possible to achieve satisfactory pixel classification performance even when using
a fraction of spectral information available for a pixel~\cite{guo:2006,maggiori:2018,drumetz:2019}.
Spectral compression is often the first step in hyperspectral image
classification pipeline~\cite{vidal:2012}.

Dimensionality reduction algorithms PCA and ICA are widely used in the
hyperspectral image analysis community for the purposes of reducing
the number of channels per pixel prior subsequent analysis steps, such
as image segmentation and pixel
classification~\cite{vidal:2012,ma:2015,ghamisi:2017,jiang:2018,ahmad:2019,
  signoroni:2019, sisodiya:2020}.  Others have employed non-linear
compression techniques, such as ICA~\cite{licciardi:2018} and Wavelet
transform~\cite{moser:2018,aroma:2020}, for compressing hyperspectral
images.  Dua \emph{et. al.} provides a survey of various compression
methods for hyperspectral images~\cite{dua:2020}.  Local Linear
Embedding (LLE)~\cite{roweis:2000,rasti:2020}, Laplacian
Eigenmaps~\cite{rasti:2020}, image quantization
techniques~\cite{dua:2020,li:2020_satellite}, and compressing a
sequence of hyperspectral images together (sometimes referred to as
\emph{temporal compression})~\cite{dua:2020} have also been used in
the hyperspectral image analysis community to reduce the amount of
data that needs to be stored and processed.

A common class of methods for ``compressing'' hyperspectral images
is~\emph{band selection}~\cite{sun:2019}.  Band selection methods
are used for a variety of analysis tasks in hyperspectral images,
including ranking, searching, clustering, constructing sparse
representations, etc.  Farrell and Mersereau studied the impact
of PCA on hyperspectral images classification when targets pixels
have similar spectral profile to those of background pixels.
They found that the PCA compression had negligible effect on the
performance of various classification methods.  This work; however,
did not study the classification performance as compression rate is varied.

As stated earlier, PCA is a commonly used compression method for
hyperspectral images.  PCA is a linear method, whereas it is
well-known that the relationship between various ``bands'' of a
spectral is highly non-linear.  There are many reasons for it,
including reflection, refraction, and the absorption property of materials
that are being imaged, plus the noise inherent in the system due to
atmospheric absorption and scattering. Cheriyadat and Bruce
demonstrated the negative effects of PCA when used as pre-processing
step for classification tasks~\cite{cheriyadat:2003}.  Du \textit{et
  al.} used ICA as a compression step for 6-band hyperspectral image
classification~\cite{du:2003}.  Here, the compressed image comprised 4
bands.  The authors noted that classification scores when using ICA with
manual band-selection were better than the classification scores
obtained when using PCA.  They also showed that the classification
performance using ICA with manual band-selection was worse than the
classification performance on the full 6-band image.

More recently, the use of autoencoder methods to compress HSI is also
increasing. Ball \textit{et al.} mentioned the use of AEs for dimensionality
reduction or for Remote Sensing datasets and not only HSI,
while Paoletti~\textit{et al.} ~\cite{paoletti:2019} described the rise
of AE-based compression as a critical pre-processing step for HSI
Hyperspectral pixels. Zhang \textit{et al.} proposed an AE model for
compression in a pipeline for unsupervised learning. However, the
previous authors usually selected the best compression rate for their tasks
and did not evaluate the variation on the results to the entire range
of compression rates.

\section{Hyperspectral datasets}
\label{sec:datasets}

\begin{figure*}
  \centerline{
    \includegraphics[width=.98\linewidth]{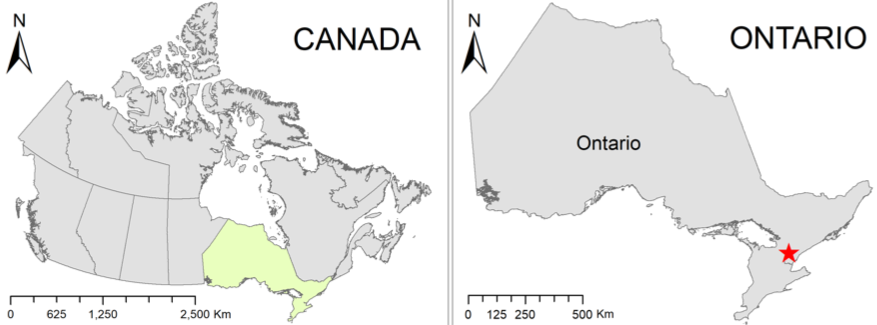}
  }
  \vspace{0.2cm}
  \centerline{
    \includegraphics[width=0.98\linewidth]{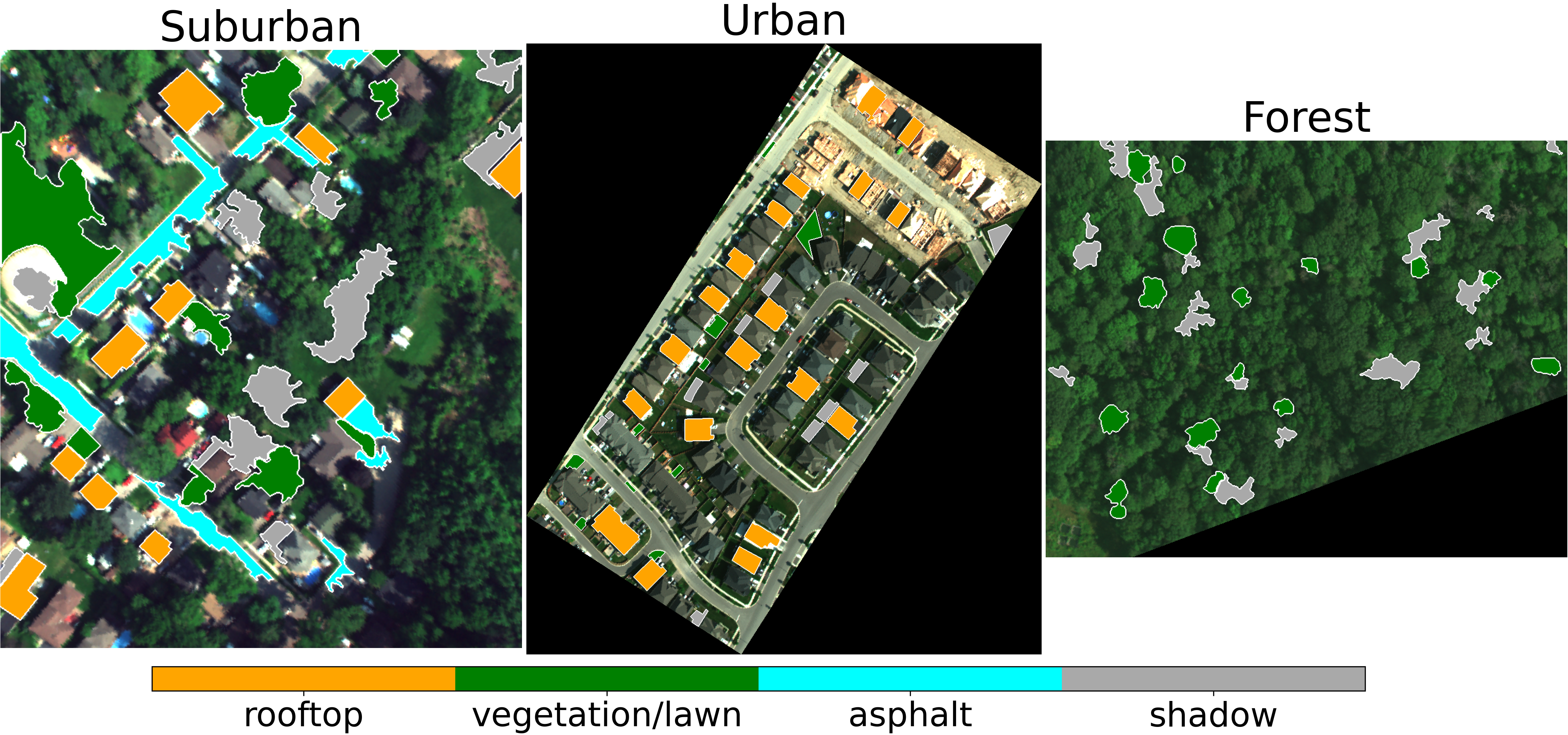}
  }
  \caption{Hyperspectral dataset was collected using an airborne
    sensor by the Remote Sensing and Spatial Ecosystem Modeling (RSSEM)
    laboratory, the Department of Geography, Geomatics and
    Environment, University of Toronto around the Toronto region
    (depicted by the red star) in Ontario, Canada.  The bottom row
    shows the three datasets in pseudo color (RGB images).  This
    visualization was constructed using the 670 nm (red), 540 nm
    (green), and 470 nm (blue) bands from original data. The yellow,
    green, blue, and gray polygons overlaid on the hyperspectral images
    are the areas with ground-truth pixel labels available.}
  \label{fig:study_area}
\end{figure*}

We used three high spatial resolution hyperspectral images for the
studies presented in this paper (Figure~\ref{fig:study_area}).  These
images were captured using the Micro-HyperSpec III sensor (from
Headwall Photonics Inc., USA) mounted at the bottom of a helicopter.
The images were captured during the daytime at 10:30 am on August 20,
2017.  The original images with 325 bands were resampled to
obtain 301 bands from 400 nm to 1000 nm with an interval of 2 nm.  Raw
images were converted to at-sensor radiance using HyperSpec III
software.  The images were also atmospherically corrected to surface
reflectance using the empirical line calibration
method~\cite{dao:2019} with field spectral reflectance measured by
FieldSpec 3 spectroradiometer from Malvern Panalytical, Malvern,
United Kingdom.  These images represent 1) urban, 2) transitional
suburban, and 3) forests landcover types. These three landcover types
cover a large fraction of use cases for hyperspectral imagery; urban
and sub-urban images are often used for city planning and land use
analysis and forest images are typically used for forest management, ecological
monitoring, and vegetation analysis.  The overlaid polygons in
Figure~\ref{fig:study_area} depict the annotated regions for which ground-truth
pixel labels are available.

\begin{table}
    \centering
    \caption{splits of train,test, and validation samples for \textbf{Suburban} dataset}
    \label{table:train_test_split_Suburban}
    \begin{tabular}{lrrr}
        \toprule
        \textbf{label}      & train & validation & test \\
        \midrule
        \textbf{Asphalt   } & 9155  & 4578       & 4578 \\
        \textbf{Rooftop   } & 7910  & 3955       & 3955 \\
        \textbf{Shadow    } & 10385 & 5192       & 5193 \\
        \textbf{Vegetation} & 15147 & 7573       & 7574 \\
        \bottomrule
    \end{tabular}
\end{table}

\begin{table}
    \centering
    \caption{splits of train,test, and validation samples for \textbf{Urban} dataset}
    \label{table:train_test_split_Urban}
    \begin{tabular}{lrrr}
        \toprule
        \textbf{label}   & train & validation & test  \\
        \midrule
        \textbf{Lawn   } & 3432  & 1716       & 1716  \\
        \textbf{Rooftop} & 22323 & 11162      & 11162 \\
        \textbf{Shadow } & 4384  & 2192       & 2192  \\
        \bottomrule
    \end{tabular}
\end{table}

\begin{table}
    \centering
    \caption{splits of train,test, and validation samples for \textbf{Forest} dataset}
    \label{table:train_test_split_Forest}
    \begin{tabular}{lrrr}
        \toprule
        \textbf{label}  & train & validation & test \\
        \midrule
        \textbf{Shadow} & 9200  & 4600       & 4600 \\
        \textbf{Tree  } & 7343  & 3672       & 3672 \\
        \bottomrule
    \end{tabular}
\end{table}

Figure~\ref{fig:study_area} (second row, left) shows the hyperspectral
image collected in an urban-rural transitional area. We refer to this
image as ``Suburban'' dataset. It was captured around Bolton area in
southern Ontario and covers an area between $43\degree 52'32"$ and
$43\degree 53'04"$ in latitude and $-79\degree 44'15"$ and $-79\degree
  43'34"$ in longitude. This region consists of various land cover
types, such as rooftops, asphalt roads, swimming pools, ponds,
grassland, shrubs, urban forest, etc. The image also contains regions
that are in shadows. The image resolution is $0.3$ square meters and
the covered area is around $41,182$ square
meters. Table~\ref{table:train_test_split_Suburban} shows the number of
samples (pixels) for different landcover types used for training and
testing.

Figure~\ref{fig:study_area} (second row, middle) shows the hyperspectral
image collected in a residential urban area, also around Bolton region
in southern Ontario.  We refer to this image as ``Urban''
dataset. It contains rooftops, under-construction residences, roads,
and lawns landcover types. The dataset also exhibits regions that are
in shadows.  This image covers the area between $43\degree 45'30"$
and $43\degree 45'43"$ in latitude and $-79\degree 50'06"$ and
$-79\degree 49'51"$ in longitude. The image resolution is $0.3$ square
meters and the area after removing background pixels is
around $59,834$ square
meters. Table~\ref{table:train_test_split_Urban} shows the number of
samples (pixels) for different landcover types used for training and
testing.

Figure~\ref{fig:study_area} (second row, right) shows the
hyperspectral dataset collected in a natural forest located at a
biological site of the University of Toronto in King City region in
southern Ontario. We refer to this dataset as ``Forest'' dataset. It
covers the area between $ 44\degree 01' 58" $ and $44\degree 02'04"$
in latitude and $-79\degree 32'06"$ and $-79\degree 31'55"$ in
longitude.  The image resolution is $0.3$ square meters and the area
after removing background pixels is around $43,084$ square meters.
Table~\ref{table:train_test_split_Forest} shows the number of
samples (pixels) for different landcover types used for training and
testing.

\section{Methodology}
\label{sec:methodology}

We used the following five methods to compress pixel spectral signal:
1) PCA, 2) KPCA, 3) ICA, 4) AE, and 5) DAE.  We also trained a gradient boosted tree model
to classify the hyperspectral image pixels given their compressed signal.
In addition, we measured the reconstruction errors by recovering
the original pixel spectra from its compressed signal.
Mathematically, say $\mathbf{x}_{i} \in \mathbb{R}^{301}$ represents a
hyperspectral pixel $i$.  We used a compression method $\mathcal{E}$ to
construct the compressed signal $\mathbf{z}_{i} =
  \mathcal{E}(\mathbf{x}_i)$, where $\mathbf{z}_{i} \in \mathbb{R}^d$
and $\mathcal{E}$ is one of the following: PCA, KPCA, ICA, AE, or
DAE. Here $1 \le  d < 301$ is a controllable parameter and lower values of
$d$ means higher compression rates.  We computed classification labels
$\mathcal{C}(\mathbf{z}_i)$ for pixel $i$ using its compressed signal,
where $\mathcal{C}$ is the gradient boosted tree classifier.  We were able
to recover the original signal $\mathbf{\hat{x}}_i$ from
$\mathbf{z}_i$ and computed the reconstruction error as $\lVert
  \mathbf{\hat{x}}_i - \mathbf{x}_i \rVert^2$.

\subsection{Compression Methods}
\label{subsection:compression_methods}

Below, we discuss the compression methods used in this paper--PCA, KPCA, and ICA--
which have been widely used as dimensionality reduction methods.
The two autoencoder models (AE and DAE) used in this study are discussed
later in the section.

\subsubsection{PCA}

PCA projects the data onto a feature space that consists of the
eigenvectors of the data covariance matrix.  Dimensionality reduction
is achieved by discarding data dimensions with low variance.  The
intuition being that data dimensions that exhibit low variance
contains little useful information. We refer the reader
to~\cite{jolliffe:2011} and~\cite{pearson:1901} for more information
on PCA.

\subsubsection{KPCA}

Kernel PCA is an extension of the PCA.  Here, input data are mapped to
a higher dimensional space using a kernel.  As per
\emph{Vapnik-Chervonenkis} theory, data mapped to a higher dimensional
space provide better separability.  Popular kernel choices are
Gaussian, Polynomial, Radial Basis Functions, and Hyperbolic Tangent.
In this work, we used a polynomial kernel, which is well-suited
to capture any non-linearities present in the data.  More information
about KPCA is available in
~\cite{cao:2003,datta:2018,liao:2010,gunter:2007}.

\subsubsection{ICA}

ICA decomposes the input signal into additive subcomponents under the
non-Gaussian and statistical independence assumptions.  It is then
possible to represent the original signal using a subset of the
independent components returned by the ICA method, thereby performing
data compression.  We refer the reader to
\cite{du:2003,comon:1994,hyvarinen:2000,stone:2004,goodfellow:2016}
for further details on ICA.

\subsubsection{AE}

We used the AE model proposed by Hinton \textit{et al.}~\cite{hinton:2006}.  It
consists of two parts: 1) an encoder, which transforms the input signal
$\mathbf{x}$ into a lower-dimensional signal $\mathbf{z}$; and 2) a
decoder, which reconstructs the original signal $\mathbf{\hat{x}}$
from the latent representation $\mathbf{z}$. Specifically, the encoder
contains of a single hidden layer, and it transforms $301$ dimensional
pixel spectra into a $d$ dimensional vector.  The decoder also consists
of a single hidden layer, and it reconstructs the $301$ dimensional signal
from a $d$ dimensional vector.  Both encoder and decoder use ReLU activation
functions for the hidden layers. The output layer of the decoder uses
the Sigmoid activation function as the expected values of the reconstructed signal are
restricted to the values of reflectance, i.e., between $0$ and $1$.
We refer the reader to~\cite{wang:2016} for technical details about our autoencoder model.
The number of elements (i.e., neurons) in the hidden layer is a hyperparameter.  We used the
grid search approach to estimate a ``good'' value for this
hyperparameter.  During hyperparameter selection we set the compression
rate equal to 99\% (i.e., $d$ was set to $4$).

\begin{figure*}
  \centerline{
    \includegraphics[width=.33\linewidth]{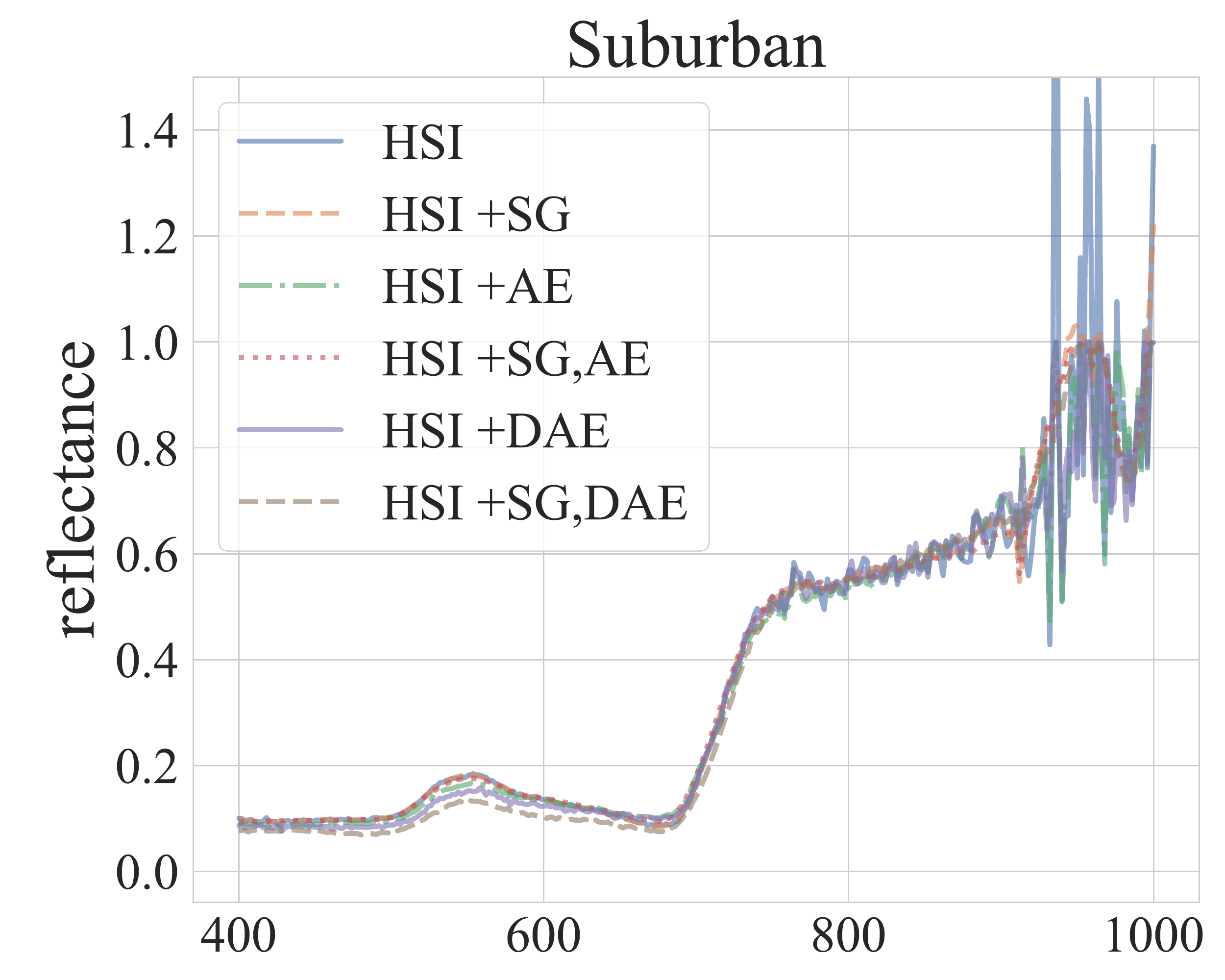}
    \includegraphics[width=.33\linewidth]{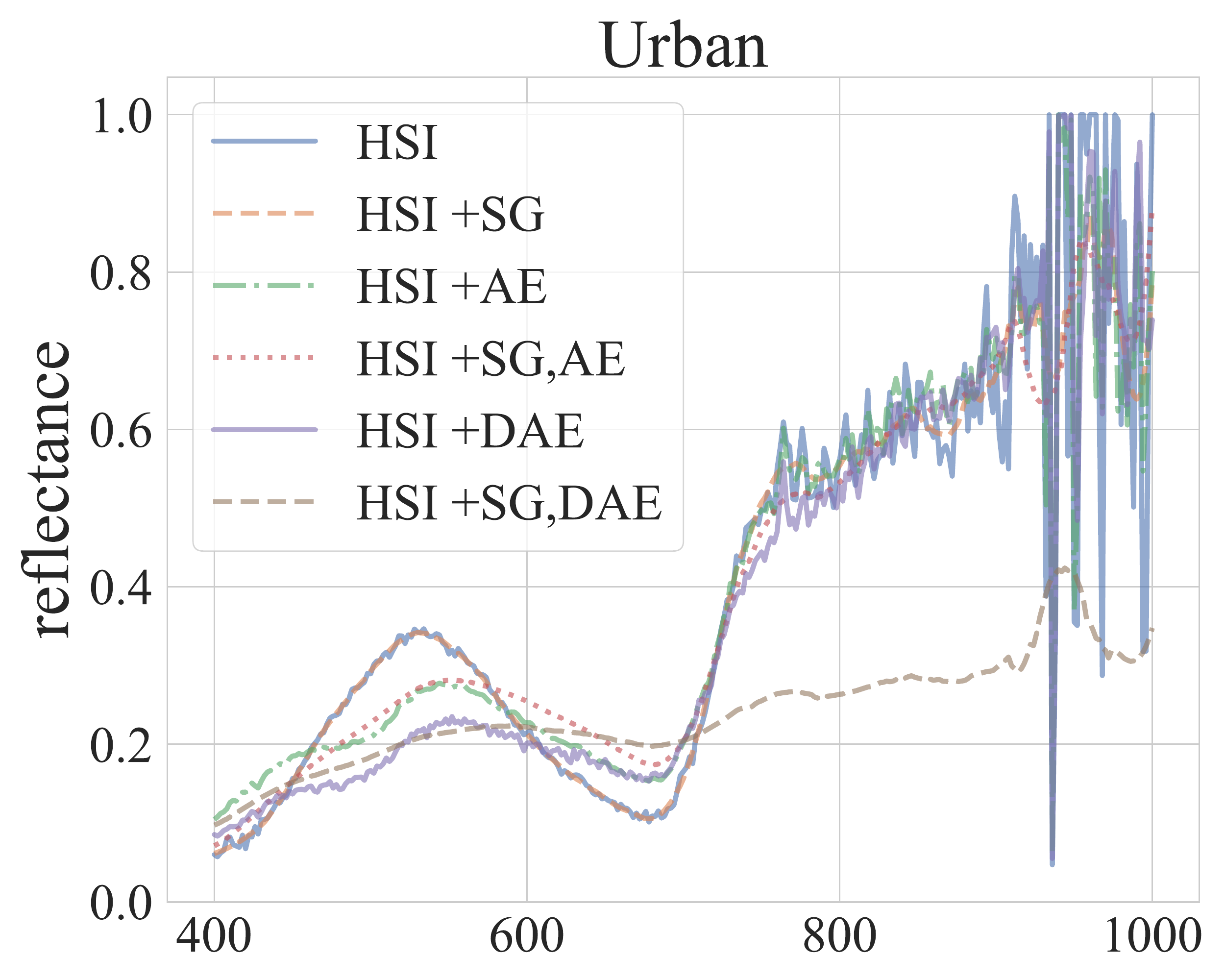}
    \includegraphics[width=.33\linewidth]{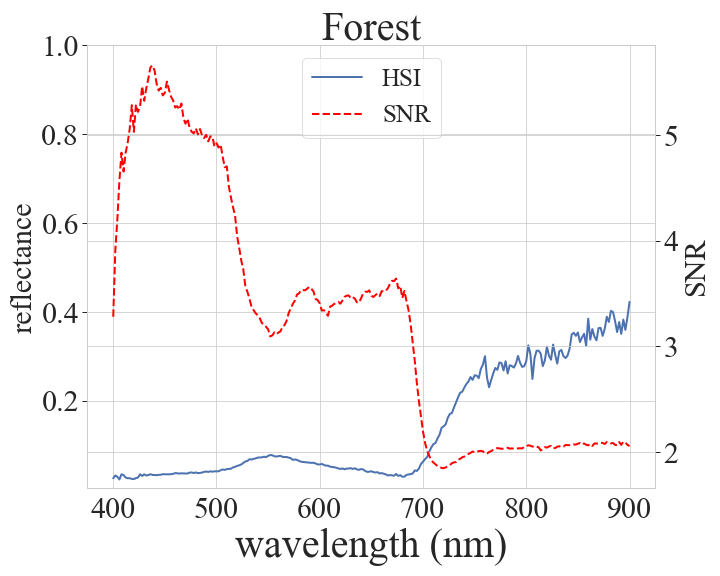}
  }
  \centerline{
    \includegraphics[width=.33\linewidth]{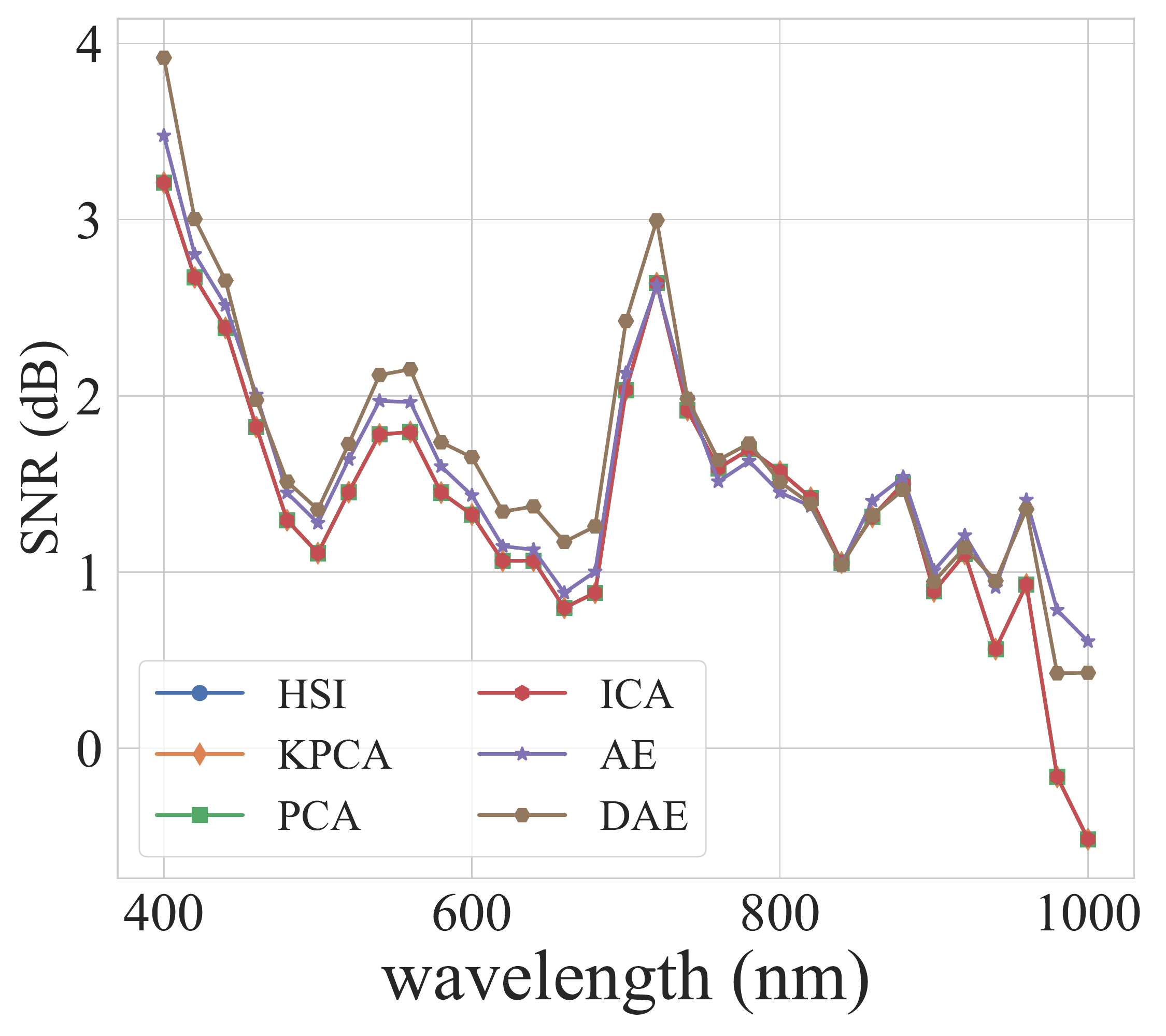}
    \includegraphics[width=.33\linewidth]{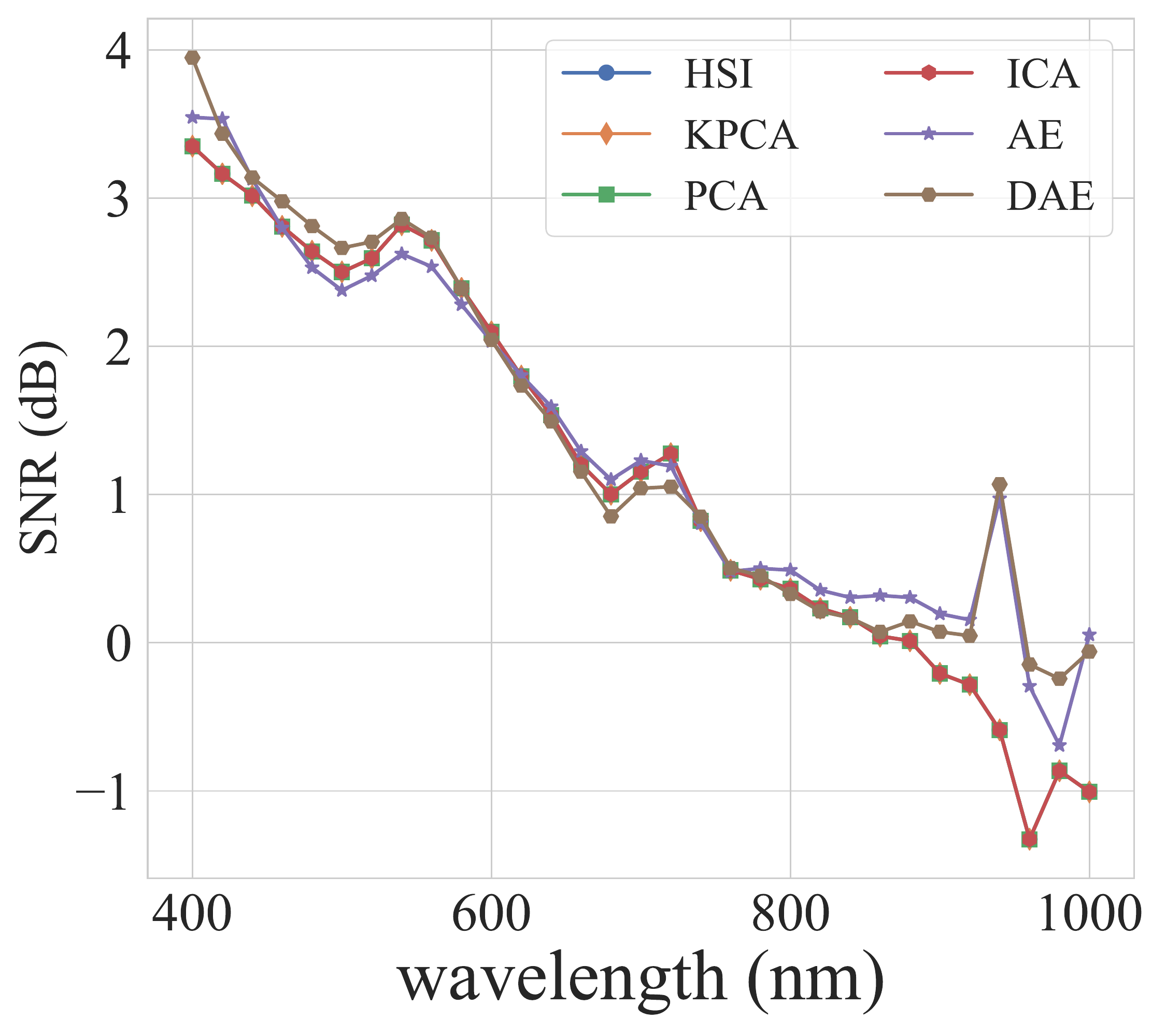}
    \includegraphics[width=.33\linewidth]{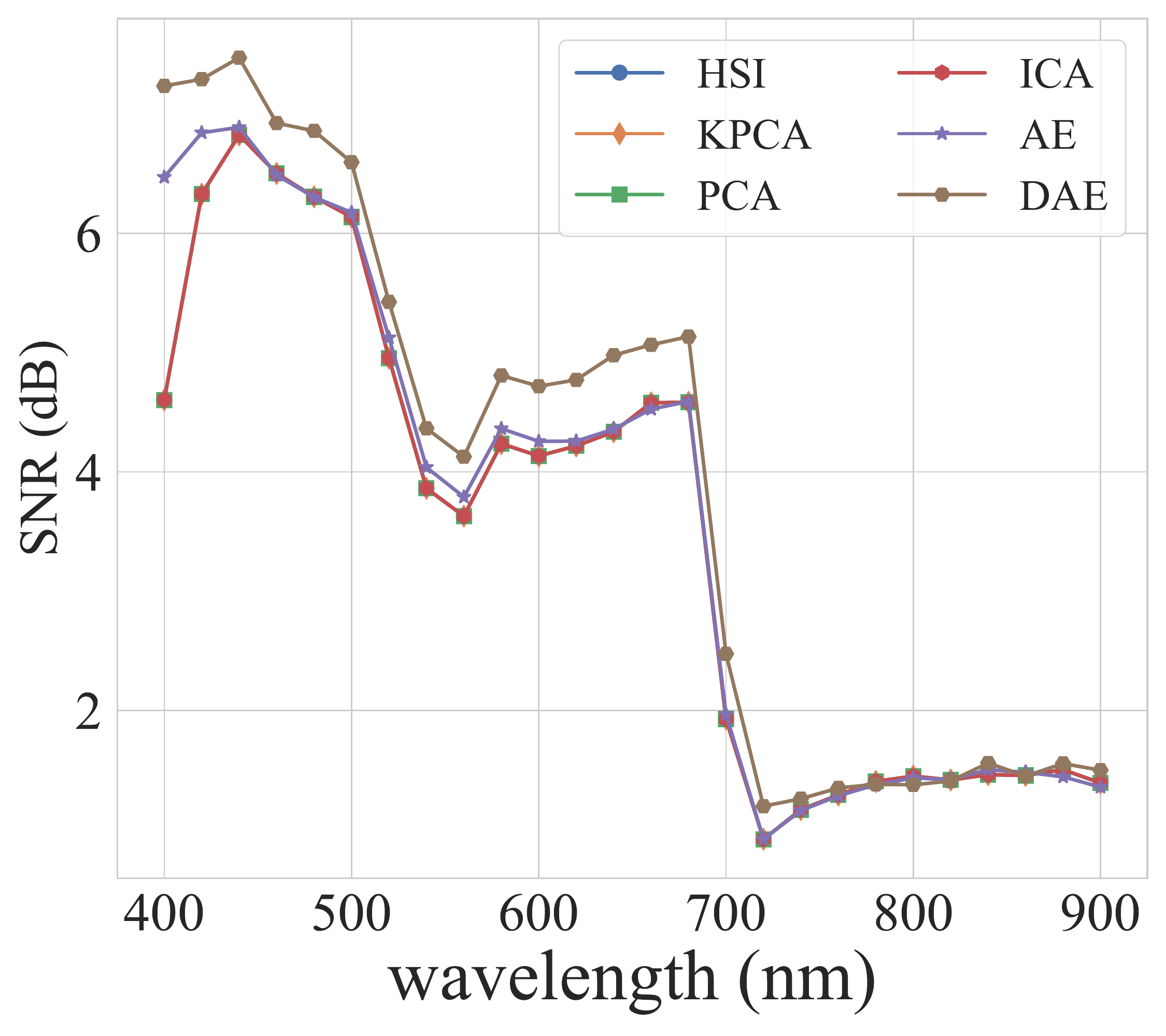}
  }
  \caption{\textbf{First row:} Spectral reconstructions for a randomly selected
    pixel in the three images. HSI denotes the original spectral
    signal.  HSI+SG refers to the denoised spectral signal.  HSI+AE and
    HSI+DAE denote reconstructed spectral signals using autoencoder
    and denoising autoencoder, respectively.  HSI+SG+AE and HSI+SG+DAE
    denote reconstructed spectral signal using transformed encodings
    (0\% Compression rates) from denoised signals (HSI+SG).
    \textbf{Second row:} Signal-to-Noise Ratio of the reconstructed spectra (PCA, KPCA, ICA, AE, DAE)
    compared to the original pixel (HSI)
  }
  \label{fig:hsi_bands_sample}
\end{figure*}

\subsubsection{Denoising AE}

It is well-known that hyperspectral images exhibit a higher degree of
noise as compared to the noise present in ordinary RGB images.
Furthermore, the level of noise present in different bands of a
hyperspectral image varies between bands.  Atmospheric water vapor,
for example, affects near-infrared bands more than higher frequency
bands.  If left untreated, noise will place an adverse effect on the
subsequent processing and analysis tasks, such as compression,
segmentation, or classification.  We implemented a denoising
autoencoder, which accounts for the noise present in the signal, for
compressing the input spectral signal~\cite{ball:2018,gondara:2018}.
Denoising autoencoder also consists of an encoder and a decoder.  The
encoder consists of two hidden layers.  The first hidden layer
contains $400$ neurons and the second hidden layer contains $500$
neurons.  The decoder also consists of two hidden layers.  The first
hidden layer contains $500$ and the second hidden layer contains $400$
neurons.  All hidden layers use ReLU activation function.  Decoder's
output layer uses Sigmoid activation function.

In order to understand the effect of noise on hyperspectral images, we
selected Saviszky-Golay (SG) algorithm, a widely used noise filtering method
for hyperspectral images, to construct clean spectral
signals~\cite{vaiphasa:2006,ruffin:1999}.
Figure~\ref{fig:hsi_bands_sample} shows spectral curves for a randomly
selected pixel in the three datasets. Figure~\ref{fig:hsi_bands_sample}
(middle) suggests that HSI+SG+DAE model did poorly in signal
reconstruction, especially in the 500–800 nm range. We observe a
similar trend for other pixels in the dataset. It appears that SG+DAE
strongly attenuates the signal in this range.  This confirms that it
is unnecessary and perhaps counter-productive to use a denoising
preprocessing step when using a denoising autoencoder to compress
hyperspectral signal. The second row of Figure~\ref{fig:hsi_bands_sample}
shows the SNR (Signal-to-Noise ratio) of each compression method compared
to the original HSI image. This result demonstrates that AE and DAE methods
can improve the SNR of the signal, suggesting that a pre-processing step
for denoising is not necessary when AE or DAE is used as a compression
algorithms.

\begin{figure}
  \centerline{
    \includegraphics[width=\linewidth]{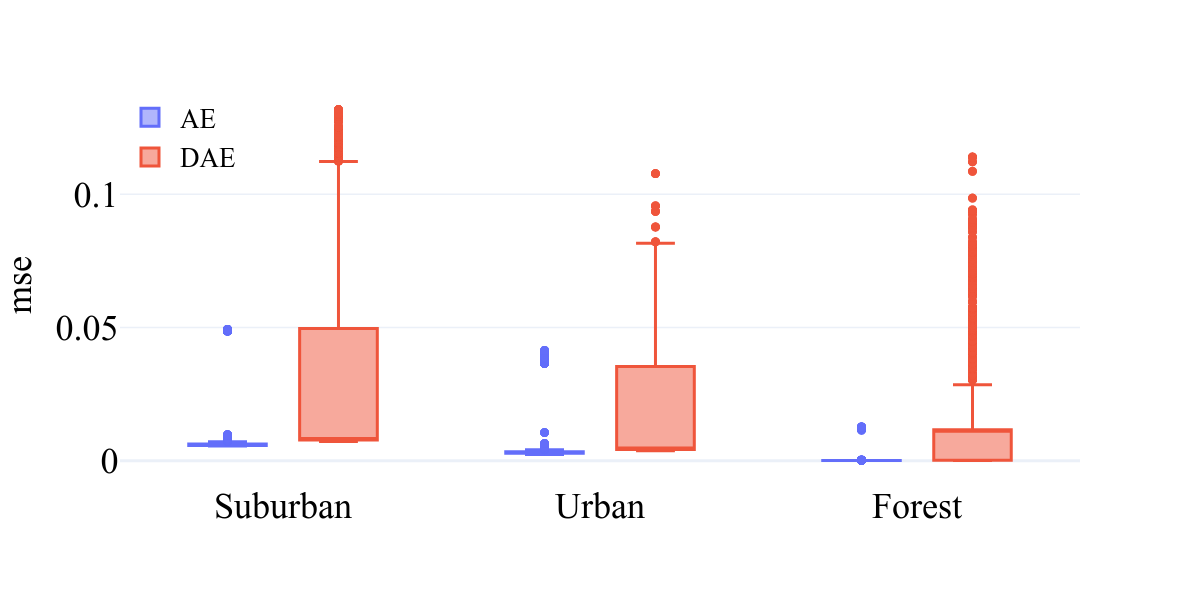}
  }
  \caption{Model variance.  Reconstruction errors for AE and DAE models for ten training runs.}
  \label{fig:AE_DAE_variance}
\end{figure}

\subsubsection{Training regime for AE and DAE}

Both autoencoder and denoising autoencoder were trained using
reconstruction loss, which is defined as $\lVert \mathbf{\hat{x}}_i -
  \mathbf{x}_i \rVert^2$.  In our experiments, both autoencoders were
able to achieve low reconstruction errors even for high compression
rates.  Tables \ref{table:train_test_split_Suburban},
\ref{table:train_test_split_Urban}, and
\ref{table:train_test_split_Forest} list the number of training and
testing samples for the suburban, urban, and forest datasets,
respectively. Each model was trained for 30 epochs using Adam
optimizer.  We trained each model ten times to capture the model
variance.  Figure~\ref{fig:AE_DAE_variance} shows reconstruction errors
for the three datasets for ten different runs for AE and DAE models.
As expected, the reconstruction errors for DAE models exhibit a larger
variance than those for AE models.  For each image we selected the
model with the lowest reconstruction error to be used as the
compression method in the final classification pipeline.

\subsection{Gradient Boosted Tree Classifier}

We employed a Gradient Boosted Tree (XGBClassifier) classifier for
pixel classification \cite{geron:2019, vasilev:2019}. XGBClassifier
is a widely used ensemble model and similar to other ensemble
methods, it avoids overfitting and offers good generalization
properties \cite{breiman:2001}. It is also easy to construct
intuitive interpretations of how this model arrives at a particular
classification decision.  We used the XGBoost library to setup our
classification model.  In our model, the number of trees was set to
$10$ and the maximum depth per tree was also set to $10$.

\subsection{Classification Metrics}

We used three metrics to evaluate the accuracy of classifications.
{\it Precision} is defined as
$$ \mathrm{\it Precision} = \frac{t_p}{(t_p + f_p)},$$
{\it recall} is defined as
$$ \mathrm{\it Recall}=\frac{t_p}{(t_p + f_n)}, $$
and {\it f1-score} is defined as the harmonic mean of precision
and recall:
$$ \text{\it f1-score} = \frac{t_p} {\left(t_p+ \frac{(f_p+f_n)}{2} \right) }. $$
Here $t_p$ is the number of true positives, $f_p$ is the number of false positives, and $f_n$ is the number of false negatives.

\section{Experiments and Results}
\label{sec:results}

\begin{figure*}
  \includegraphics[width=\linewidth]{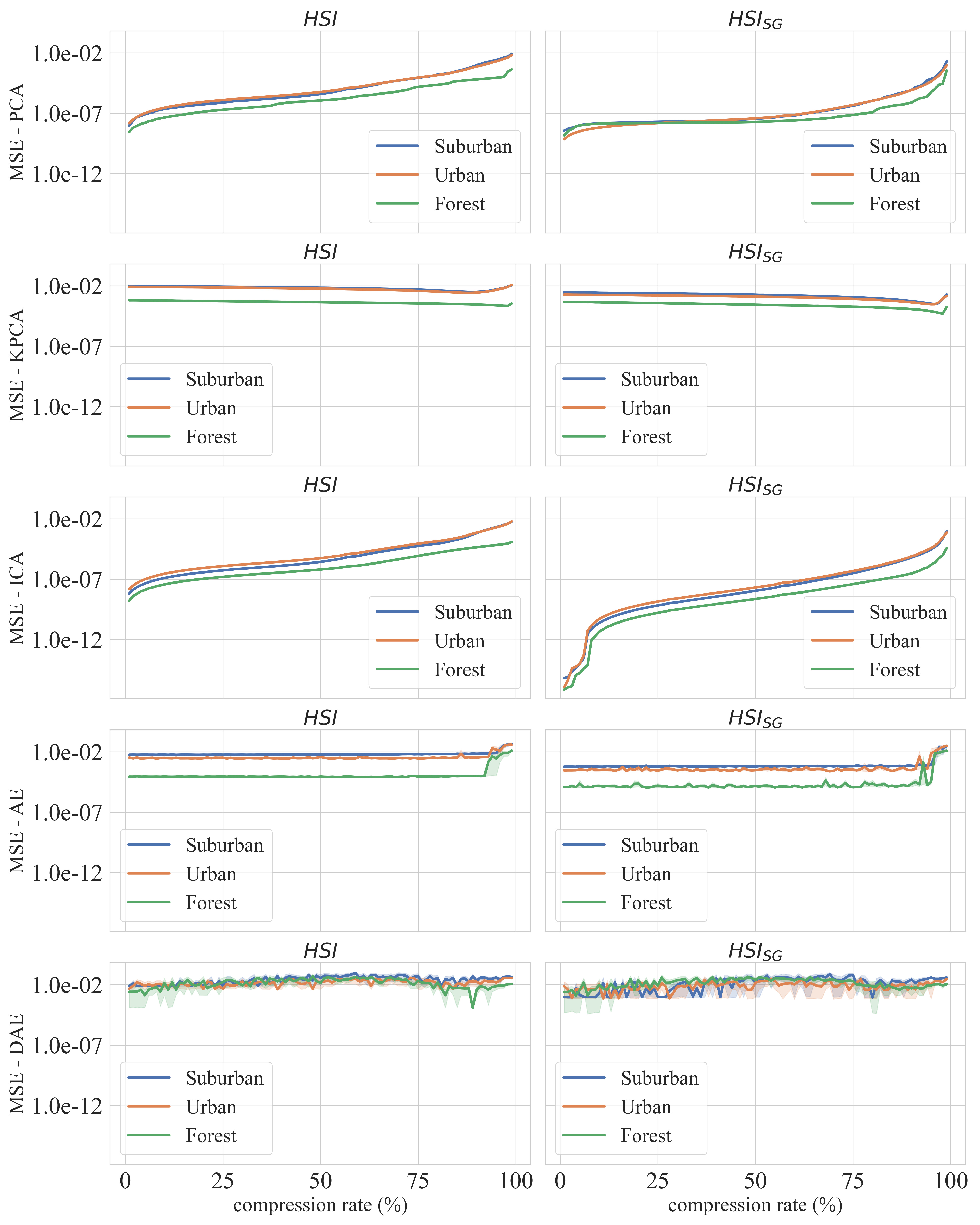}
  \caption{MSE errors for the 3 datasets and 5 compression
    algorithms, from top to bottom: a) Principal Component Analysis,
    b) Independent Component Analysis, c) Kernel Principal Component
    Analysis, d) AutoEncoder, e) Denoising AutoEncoder}
  \label{fig:mse_reconstruction}
\end{figure*}

A standard way to study the performance of different compression
algorithms is to recover the original signal from its compressed
version as depicted in Figure~\ref{fig:mse_reconstruction}. In the
following sections, we examine reconstruction errors for PCA, KPCA,
ICA, AE and DAE for different compression rates.  We also present
reconstruction errors both with and without the SG noise reduction
pre-processing step. As stated earlier, compressing hyperspectral data
is desirable; however, we are also interested in pixel-level classification
using the compressed data.  We define pixel-based classification as the problem of
identifying landcover type, say forest, rooftop, etc., for a given
pixel in an hyperspectral image. Within this context, we seek the
answer to the following two questions: a) how compression rates affect
pixel classification scores and b) for a given compression rate, which
compression method achieves the highest classification accuracy.

\subsection{Spectral Reconstruction}

\begin{figure*}
  \centerline{
    \includegraphics[width=.33\linewidth]{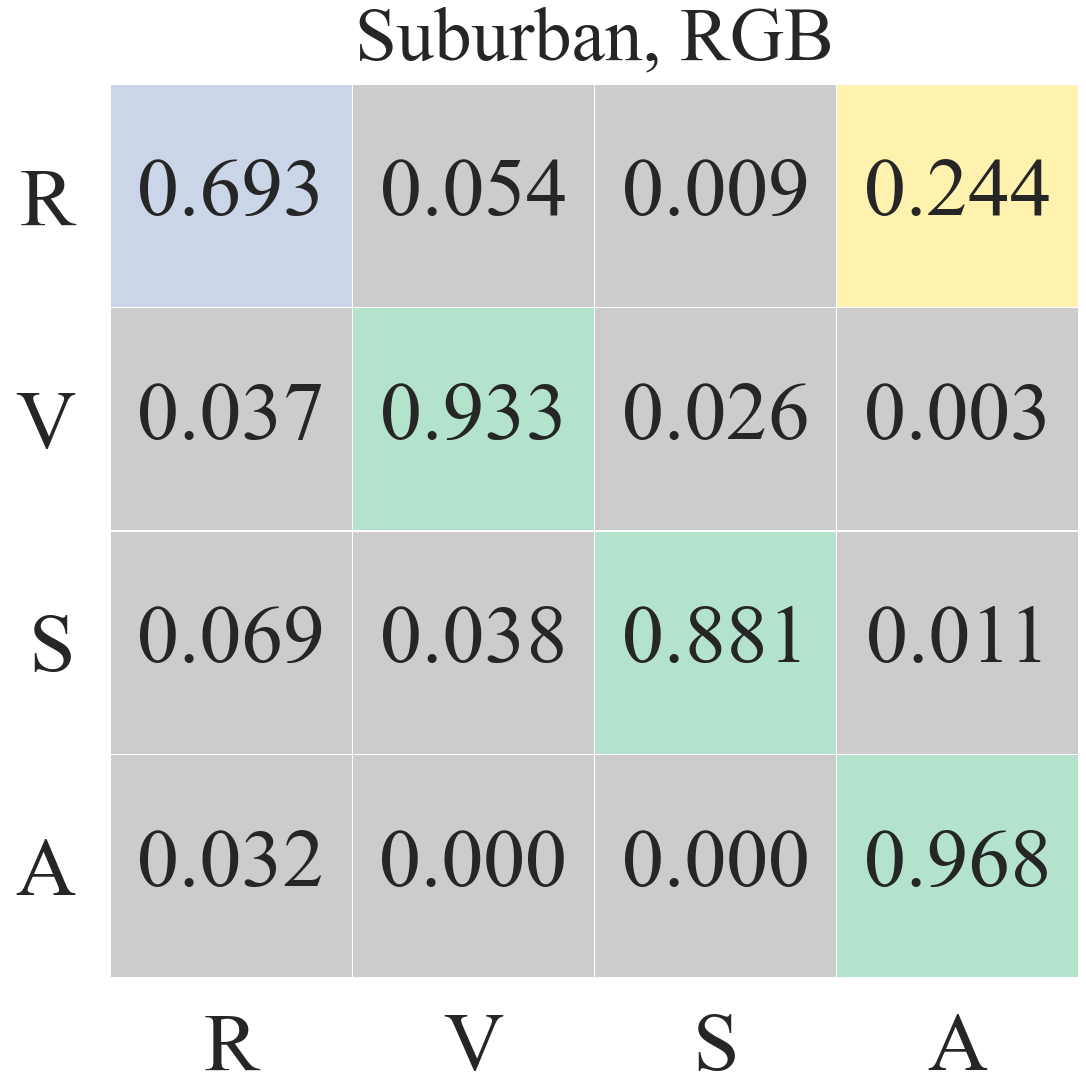}
    \includegraphics[width=.33\linewidth]{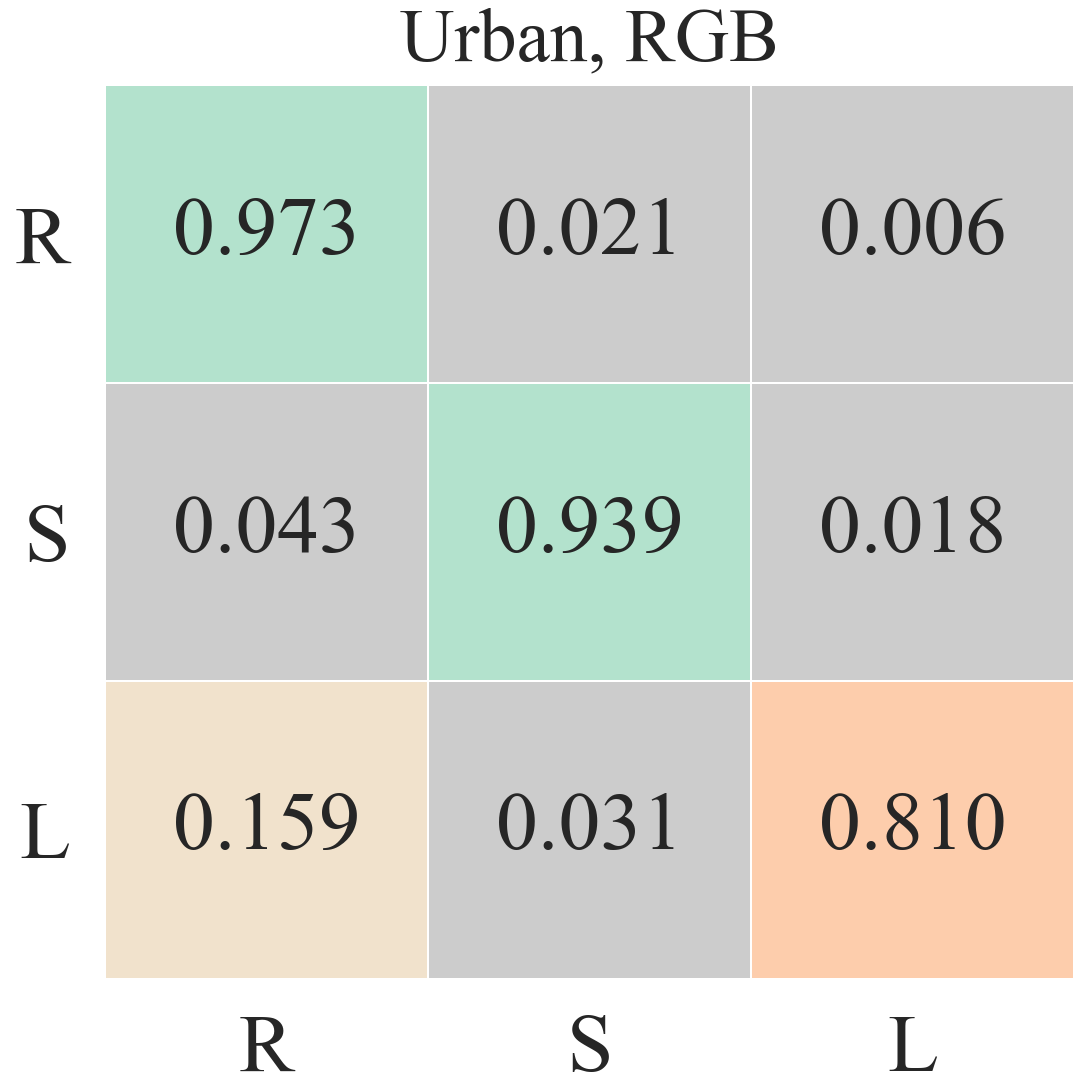}
    \includegraphics[width=.33\linewidth]{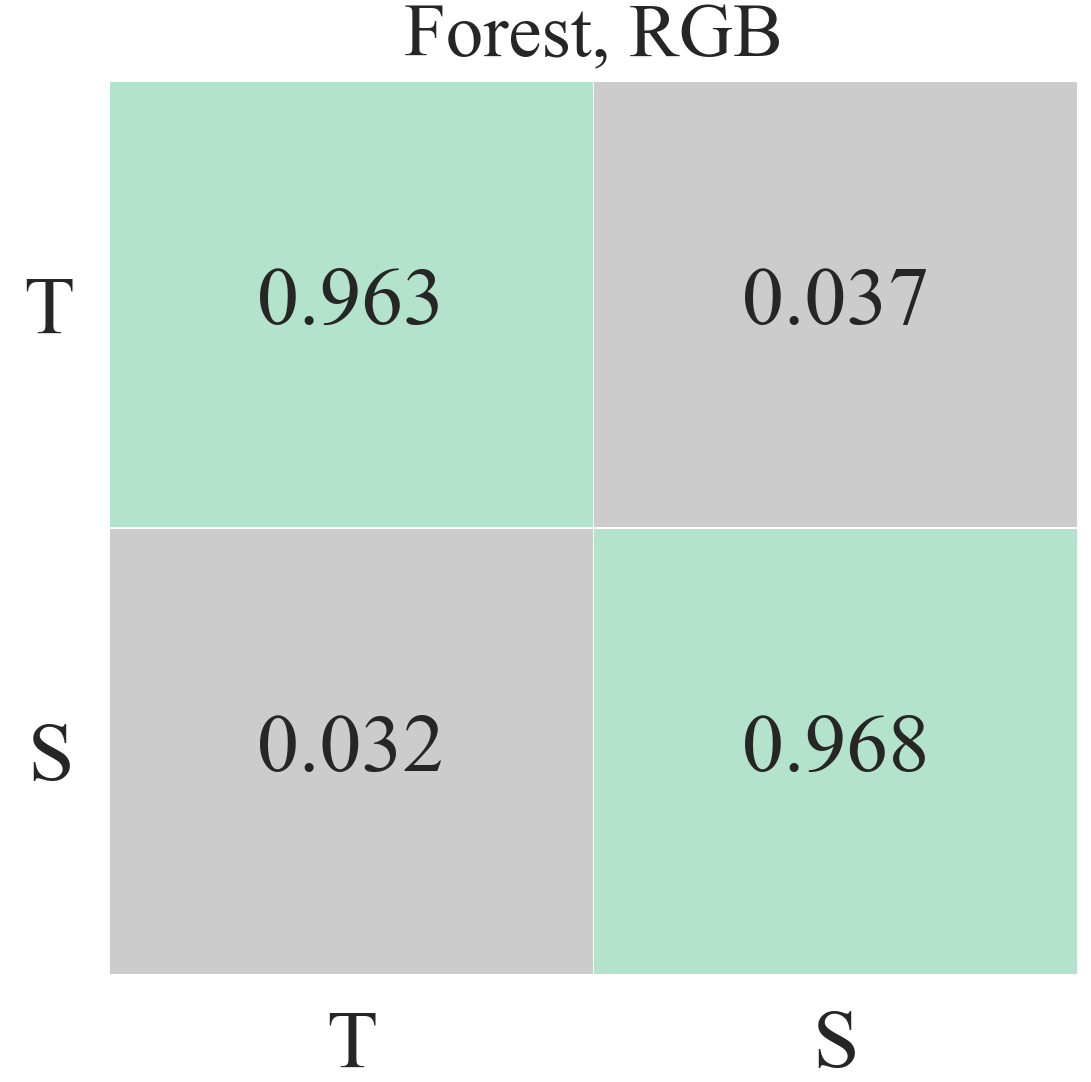}
  }
  \caption{Confusion matrix for classification scores for three
    datasets using RGB data.  (Left) R, V, S, and A refer to Rooftop,
    Vegetation, Shadow, and Asphalt; (Center) R, S, L refer to
    Rooftop, Shadow and Lawn, respectively; and (Right) T, S refer to
    Tree and Shadow, respectively}
  \label{fig:confusion_matrix_baseline_RGB}
\end{figure*}

Figure~\ref{fig:mse_reconstruction} presents reconstruction losses for
different methods for the three datasets. Here compression rates vary
from $1\%$ to $99\%$.  For our purposes, the compression rate is
defined as the ratio of $(n-d)$ to $n$, where $n$ is the number of
dimensions of the original signal and $d$ is the number of dimensions
of the compressed signal.  Recall that $n$ is equal to $301$ for the
datasets used in this paper.  Compression rates are related
to the memory needed to store the compressed data.  The left column of
Figure~\ref{fig:mse_reconstruction} shows the results for the original
hyperspectral data (HSI), whereas the right column shows the results
for the data that have been pre-processed using the SG filter
(HSI+SG).

The reconstruction errors rise as compression rates increase for both
PCA and ICA models. Notice, however, other methods---KPCA, AE, and DAE---are
able to achieve low reconstruction errors even for high compression
rates.  This effect can be explained by the fact that
non-linear methods are able to better handle non-linearities present
in the data while PCA and ICA are linear methods.  It is interesting to
note that PCA, KPCA, and ICA methods outperform deep learning methods
AE and DAE for compression rates less than ninety
percent. Furthermore, AE and DAE match the reconstruction performance
of PCA, KPCA, and ICA only for compression rates higher than ninety
percent.

The difference between reconstruction errors for original data (HSI)
and for data preprocessed using SG filter (HSI+SG) falls as
compression rate increases.  This is noteworthy since it suggests that
compression may have a denoising effect on the original spectral signal.
Curiously, we also observe a slightly higher variance in reconstruction
score for DAE method for pre-processed data (HSI+SG), which merits further
investigation, and we leave it as future work.  In the following section,
we do not apply SG to the original spectra before compression and classification,
as it does not show effective as discussed in the Section~\ref{subsection:compression_methods}

\subsection{Classification}

\begin{figure*}
  \includegraphics[width=\linewidth]{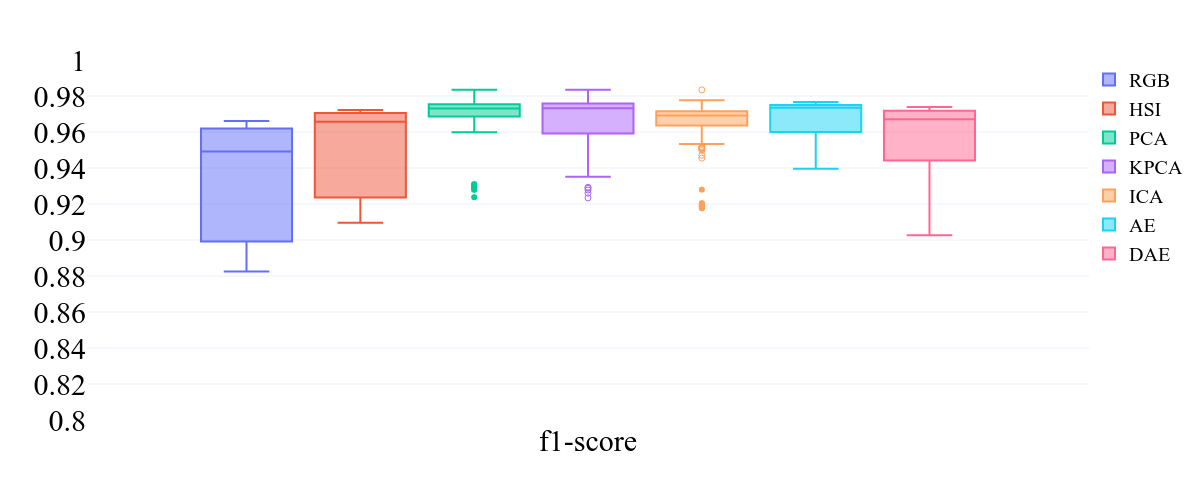}
  \caption{F1-scores across all compression rates for all datasets and
    landcover types using PCA, KPCA, ICA AE and DAE methods.
    This figure also includes precision, recall and f1 score for
    all datasets and landcover types when using RGB data for pixel
    classification}
  \label{fig:overall_scores_boxplot}
\end{figure*}

\begin{figure*}
  \centerline{
    \includegraphics[width=.33\linewidth]{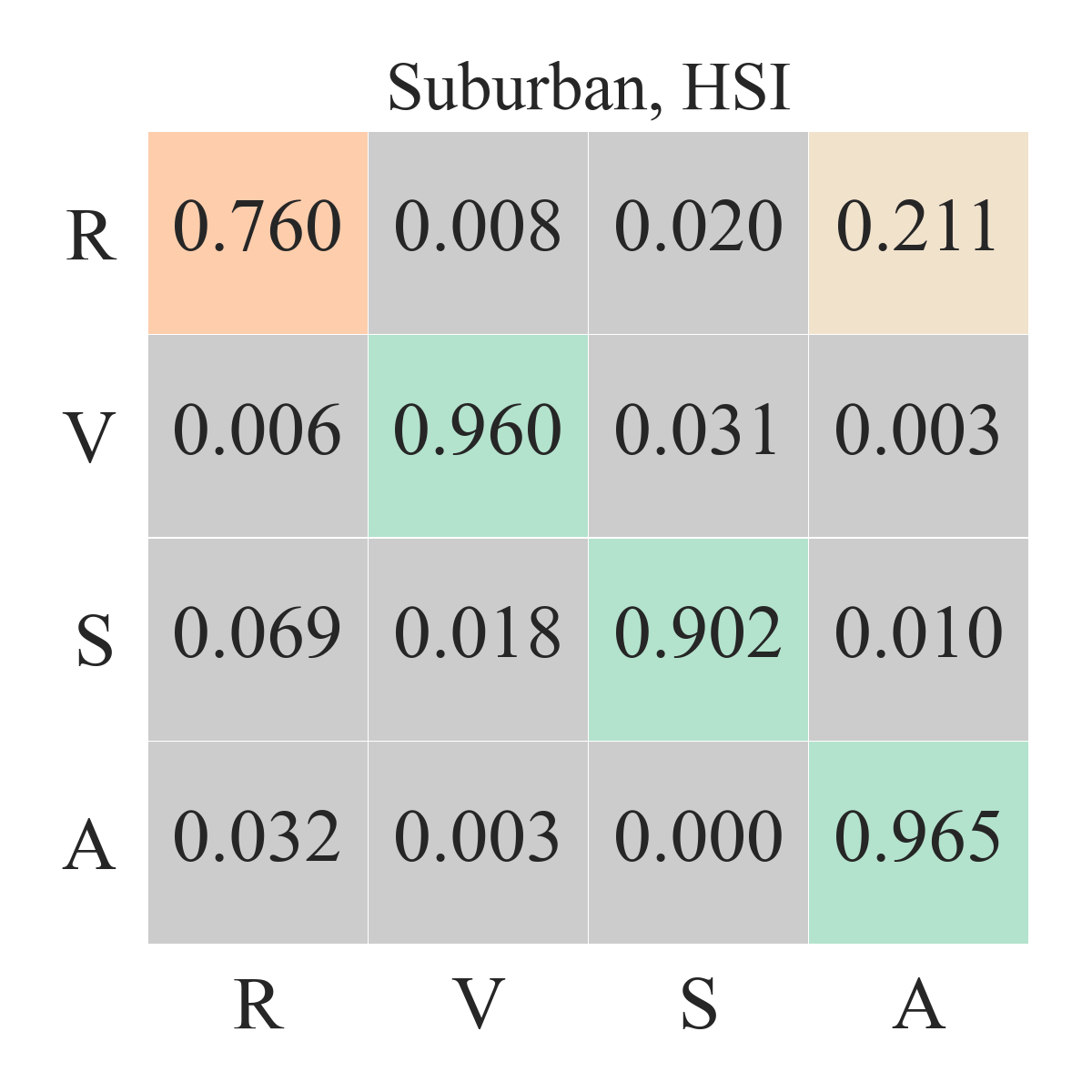}
    \includegraphics[width=.33\linewidth]{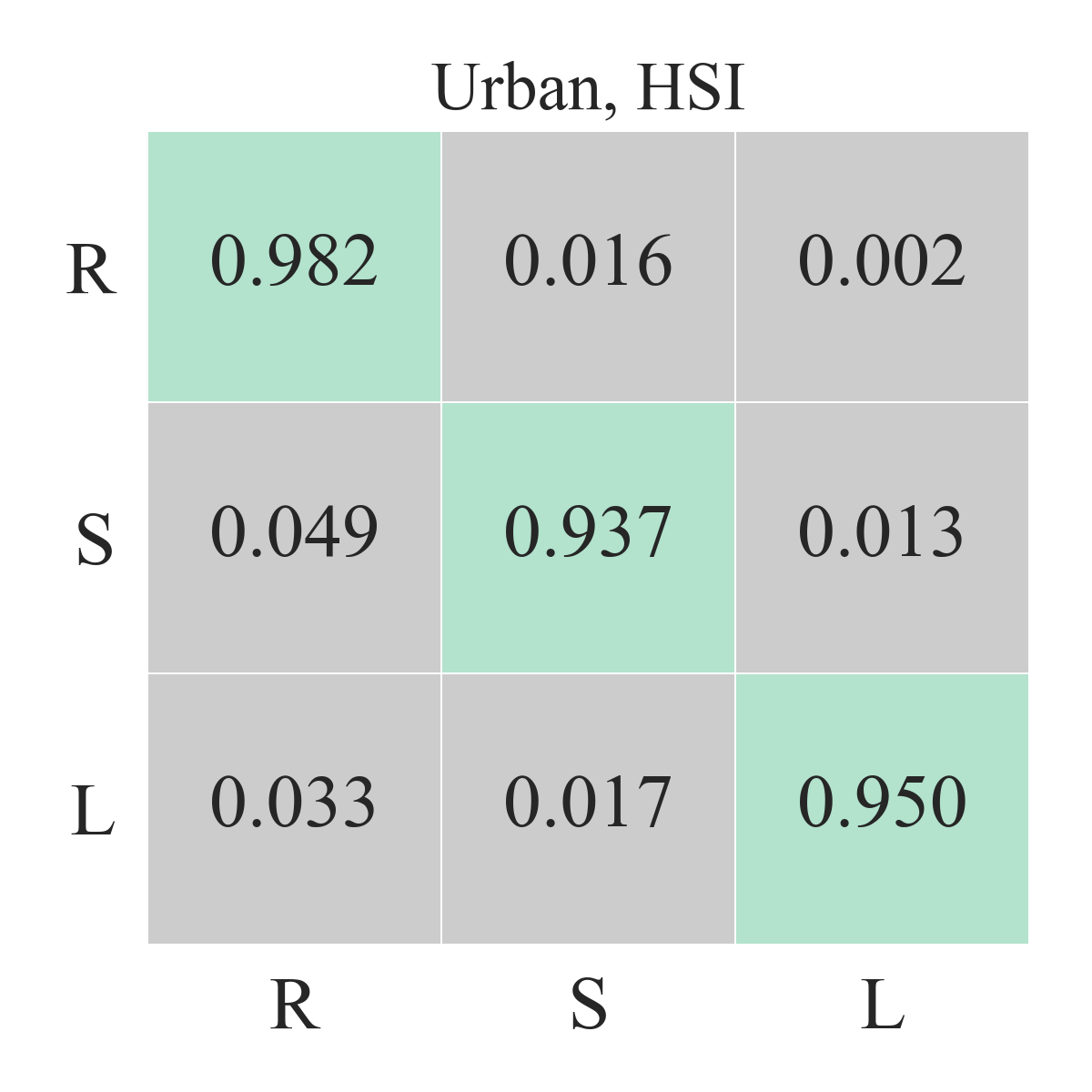}
    \includegraphics[width=.33\linewidth]{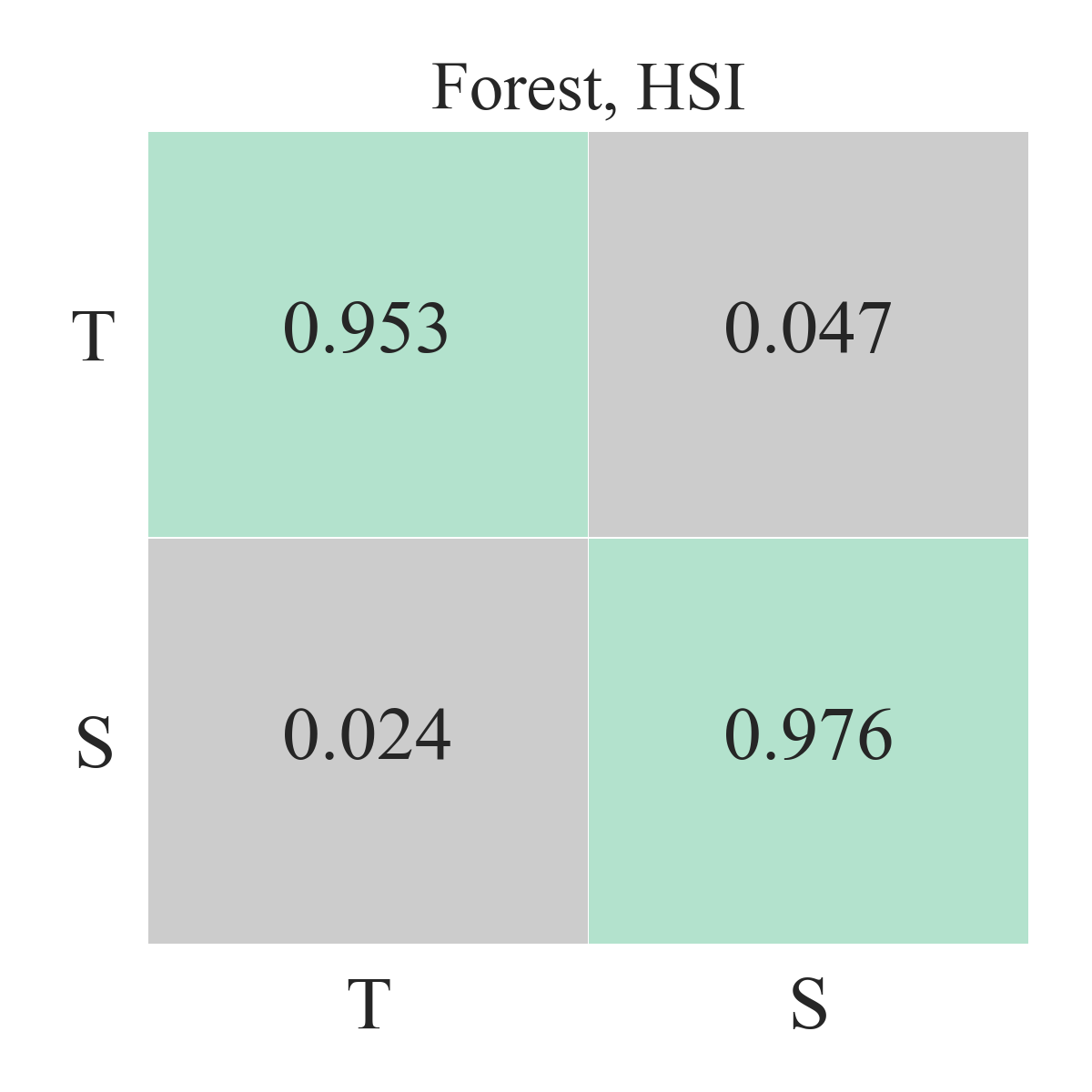}
  }
  \caption{Confusion matrix for classification scores for three datasets
    using HSI data.
    (Left) R, V, S, and A refer to Rooftop, Vegetation, Shadow, and Asphalt;
    (Center) R, S, L refer to Rooftop, Shadow and Lawn respectively; and
    (Right) T, S refer to Tree and Shadow.}
  \label{fig:confusion_matrix_baseline_HSI}
\end{figure*}

The reconstruction error is a measure of how much information is
preserved in the compressed signal, since this information is needed
to reconstruct the original signal.  The reconstruction error,
however, is not a robust measure of classification performance on the
compressed signal.  In this section we study classification
performance on compressed signal for PCA, KPCA, ICA, AE and DAE
compression methods and for different compression rates.

Figure~\ref{fig:confusion_matrix_baseline_RGB} shows the confusion
matrices for landcover classification for RGB data.  Here bands
corresponding to red $(670 nm)$, green $(540 nm)$, and blue $(470 nm)$
wavelengths are selected to form RGB pixels.  These scores provide a
baseline for the classification results obtained by using the hyperspectral
data. Figure~\ref{fig:overall_scores_boxplot} shows {\it f1-scores},
precision, and recall values obtained using 1) RGB, 2) uncompressed,
and 3) compressed hyperspectral data.  The results shown for
compressed data are aggregated over all compression rates. RGB {\it
    f1-scores} range between $0.9$ and $0.96$; however, {\it f1-scores}
obtained by using hyperspectral data fall between $0.96$ and $0.98$.
Figure~\ref{fig:confusion_matrix_baseline_HSI} shows \emph{f1-scores}, and
it confirms our intuition that the classification results obtained by using
hyperspectral data are better than those obtained by using RGB channels.
We will return to this later in this section.

\begin{figure*}
  \includegraphics[width=\linewidth]{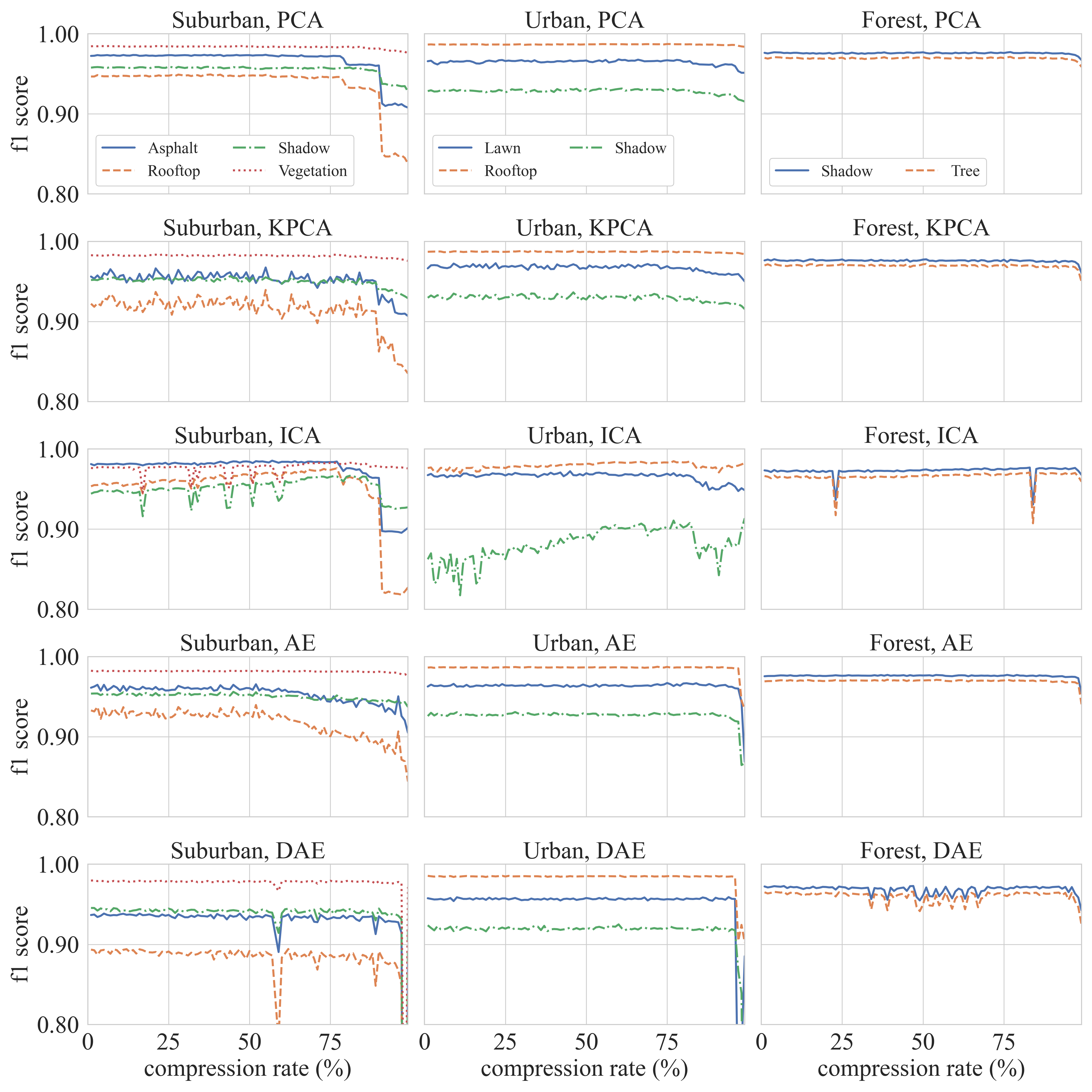}
  \caption{Classification f1 scores (using compressed data) vs. compression rates.
    First column plot results for the Suburban dataset,
    the second column plot results for the Urban dataset,
    and the last column plot results for the Forest dataset.
    The rows group the classification algorithms.
    The \emph{f1-scores} are plotted for each label present in the dataset
  }
  \label{fig:f1_score_curves_HSI}
\end{figure*}

We now turn our attention to the case when classification is performed
on compressed hyperspectral data.
Figure~\ref{fig:f1_score_curves_HSI} plots \emph{f1-scores} for PCA,
KPCA, ICA, AE and DAE compression methods vs. compression rates.  In
each case an XGBoost classifier is used to predict pixel
landcover-types.  A total of 1470  classifiers is trained, one for
each compression rate (98) for every compression method (5) and for each
dataset (3), in order to ensure that the differences in classification
scores can be explained by the ability of the compression algorithm to
encode the relevant information. Each XGBoost classifier is
trained using identical meta parameters and training regimes. These
experiments then provide a different lens for studying compression
algorithms.  Specifically, these experiments help us pose the
question: is it true that compression algorithms that achieve low
reconstruction errors also create a compressed signal that encodes the
information necessary to perform pixel-level classifications?

Ideally, we want classification algorithms that operate in the
compressed signal domain.  It is both computation and space
inefficient to have to reconstruct the original signal to perform
classification. As expected, classification performance as measured by
\emph{f1-scores} drops as compression rates increase.  At the same
time, however, nearly all methods post \emph{f1-scores} greater than
$0.85$ even for compression rates greater than eighty percent. This
suggests that it is possible to achieve good classification
performance when using a compressed hyperspectral signal.

\subsubsection{Classification using RGB Data}

The RGB classifier only achieves an accuracy of 69\% for rooftop
landcover type in the suburban dataset. Using hyperspectral data
improves upon the classification scores obtained by using RGB data.
These results confirm that \emph{f1-scores} for hyperspectral datasets
improve upon those for RGB data by around one to two percent.  Note
also that this improvement is maintained when performing
classification using the compressed data.  Specifically, our results
suggest that this improvement holds even at 98\% compression rates.
At 98\% compress rate, each hyperspectral pixel is encoded in a
6-dimensional vector, which is only twice the number that is needed to
store an RGB pixel.  We believe that classification scores using
hyperspectral data, compressed or otherwise, will pull ahead of the
scores obtained by RGB data as the number of landcover types (or
labels) increases.  We currently do not have access to a dataset that
is needed to study this issue further.

\begin{figure*}
  \centerline{
    \includegraphics[width=\linewidth]{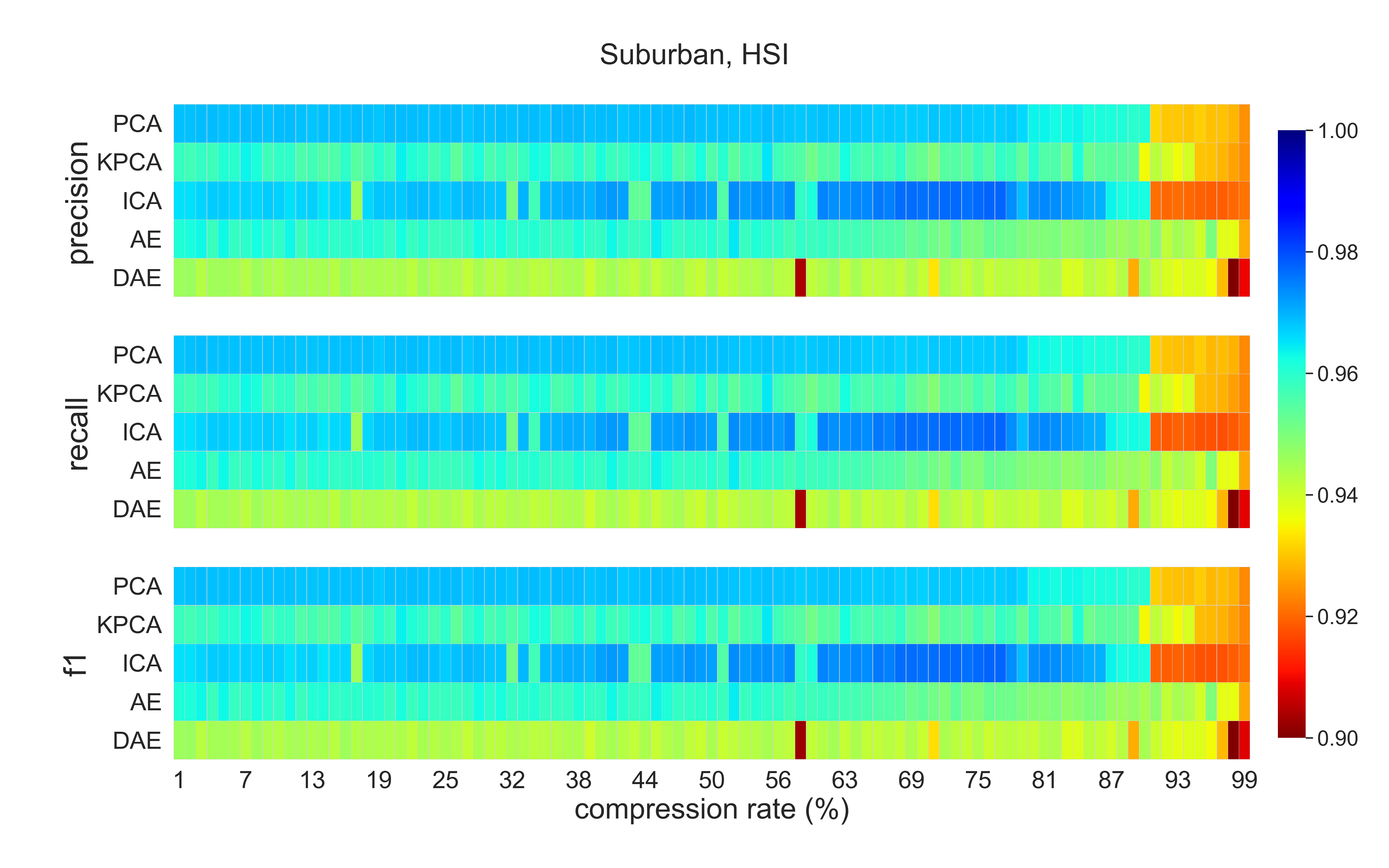}
  }
  \caption{Suburban dataset classification scores for all methods for compression rates between 1\% to 99\%.}
  \label{fig:heatmap_Suburban_HSI}
\end{figure*}

\subsection{Classification on compressed data (Suburban Dataset)}

Figure~\ref{fig:heatmap_Suburban_HSI} visualizes classification
\emph{f1-score}, recall, and precision values for compression rates
between 1\% and 99\% on Suburban dataset.  As expected scores for PCA,
KPCA, AE, and DAE compression methods decrease as the compression rate
increases.  ICA is an outlier.  ICA has higher classification
performance for compression rates between 63\% and 77\%.  PCA
compression achieves best classification performance for compression
rates less than 90\%.  AE compression method achieves best
classification performance when compression rate is between 95\% and
97\%.  While classification performance of AE and DAE compression
methods is similar to other methods for low compression rates, AE and
DAE achieve better classification performance as compared to those
obtained by other methods for compression rates between 90\% to 95\%.
Classification accuracy plummets for compression rates greater than
97\%.  Table~\ref{table:top_classification_scores_Suburban_HSI} shows
\emph{f1-scores}, precision, and recall for Suburban dataset at 95\%
compression (the compressed signal is a 15-dimensional vector, down
from 301-dimensional original spectral signal).  It also includes
these scores for the RGB data.  AE and DAE methods outperform other
methods at this compression rate.

\begin{table}[H]
    \centering \footnotesize
    \caption{Top classification scores Suburban, HSI, compression rate=95\%}
    \label{table:top_classification_scores_Suburban_HSI}
    \begin{tabular}{llrrr}
        \toprule
                                             &                      & precision & recall & f1-score \\
        \textbf{label}                       & \textbf{compression} &           &        &          \\
        \midrule
        \multirow{6}{*}{\textbf{Asphalt}}    & \textbf{RGB}         & 0.809     & 0.968  & 0.881    \\
                                             & \textbf{PCA}         & 0.888     & 0.939  & 0.913    \\
                                             & \textbf{KPCA}        & 0.885     & 0.939  & 0.911    \\
                                             & \textbf{ICA}         & 0.862     & 0.934  & 0.897    \\
                                             & \textbf{AE}          & 0.917     & 0.950  & 0.928    \\
                                             & \textbf{DAE}         & 0.914     & 0.943  & 0.928    \\
        \cline{1-5}
        \multirow{6}{*}{\textbf{Rooftop}}    & \textbf{RGB}         & 0.776     & 0.693  & 0.732    \\
                                             & \textbf{PCA}         & 0.856     & 0.845  & 0.851    \\
                                             & \textbf{KPCA}        & 0.855     & 0.842  & 0.849    \\
                                             & \textbf{ICA}         & 0.823     & 0.816  & 0.819    \\
                                             & \textbf{AE}          & 0.881     & 0.887  & 0.878    \\
                                             & \textbf{DAE}         & 0.873     & 0.880  & 0.877    \\
        \cline{1-5}
        \multirow{6}{*}{\textbf{Shadow}}     & \textbf{RGB}         & 0.952     & 0.881  & 0.915    \\
                                             & \textbf{PCA}         & 0.955     & 0.918  & 0.936    \\
                                             & \textbf{KPCA}        & 0.954     & 0.917  & 0.935    \\
                                             & \textbf{ICA}         & 0.956     & 0.896  & 0.925    \\
                                             & \textbf{AE}          & 0.960     & 0.928  & 0.943    \\
                                             & \textbf{DAE}         & 1.000     & 0.923  & 0.938    \\
        \cline{1-5}
        \multirow{6}{*}{\textbf{Vegetation}} & \textbf{RGB}         & 0.945     & 0.933  & 0.939    \\
                                             & \textbf{PCA}         & 0.980     & 0.978  & 0.979    \\
                                             & \textbf{KPCA}        & 0.979     & 0.977  & 0.978    \\
                                             & \textbf{ICA}         & 0.978     & 0.975  & 0.977    \\
                                             & \textbf{AE}          & 0.981     & 0.980  & 0.980    \\
                                             & \textbf{DAE}         & 0.977     & 1.000  & 0.977    \\
        \bottomrule
    \end{tabular}
\end{table}

\begin{figure*}
  \centerline{
    \includegraphics[width=\linewidth]{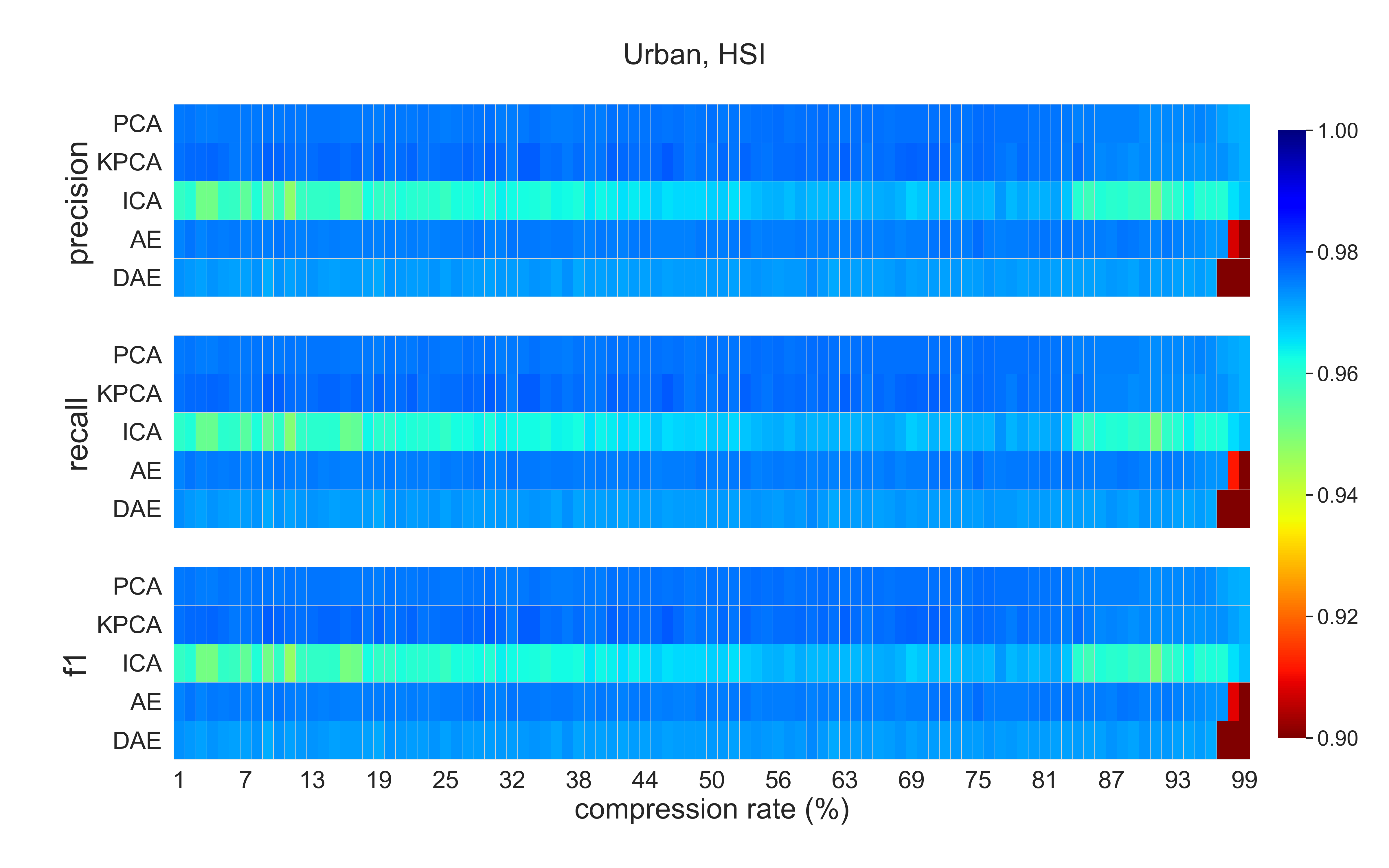}
  }
  \caption{Urban dataset classification scores for all methods for compression rates between 1\% to 99\%.}
  \label{fig:heatmap_Urban_HSI}
\end{figure*}

\subsection{Classification on compressed data (Urban Dataset)}
Figure~\ref{fig:heatmap_Urban_HSI} visualizes classification
\emph{f1-score}, recall, and precision values for compression rates
between 1\% and 99\% on Urban dataset.  PCA and KPCA are the best
performing methods; however, AE is able to match the performance of
these methods for compression rates less than 96\%. ICA method seems
to be struggling with this dataset. The performance of the DAE method
is inconsistent across the compression rates range. This can be
attributed to the stochastic nature of this method.  The
classification accuracy for data compressed using ICA is higher for
compression rates between 63\% and 77\%.  For compression rates of
less than 90\%, best classification scores are achieved when data are
compressed using PCA. AE compression method achieves the best
classification scores when the compression rate lies between 95\% and
97\%.  Table~\ref{table:top_classification_scores_Urban_HSI} shows
\emph{f1-score}, precision, and recall for the urban data at the 95\%
compression.  The table also includes these scores for the RGB
baseline. AE and DAE methods outperform other methods at this level of
compression.

\begin{table}[H]
\centering \footnotesize
\caption{Top classification scores Urban, HSI, compression rate=95\%}
\label{table:top_classification_scores_Urban_HSI}
\begin{tabular}{llrrr}
\toprule
       &     &  precision &  recall &  f1-score \\
\textbf{label} & \textbf{compression} &            &         &           \\
\midrule
\multirow{6}{*}{\textbf{Lawn}} & \textbf{RGB} &      0.930 &   0.810 &     0.866 \\
       & \textbf{PCA} &      0.963 &   0.960 &     0.962 \\
       & \textbf{KPCA} &      0.966 &   0.951 &     0.958 \\
       & \textbf{ICA} &      0.965 &   0.947 &     0.956 \\
       & \textbf{AE} &      0.967 &   0.957 &     0.962 \\
       & \textbf{DAE} &      0.961 &   0.954 &     0.957 \\
\cline{1-5}
\multirow{6}{*}{\textbf{Rooftop}} & \textbf{RGB} &      0.967 &   0.973 &     0.970 \\
       & \textbf{PCA} &      0.984 &   0.988 &     0.986 \\
       & \textbf{KPCA} &      0.983 &   0.987 &     0.985 \\
       & \textbf{ICA} &      0.971 &   0.983 &     0.977 \\
       & \textbf{AE} &      0.984 &   0.988 &     0.986 \\
       & \textbf{DAE} &      0.983 &   1.000 &     0.985 \\
\cline{1-5}
\multirow{6}{*}{\textbf{Shadow}} & \textbf{RGB} &      0.877 &   0.939 &     0.907 \\
       & \textbf{PCA} &      0.935 &   0.915 &     0.924 \\
       & \textbf{KPCA} &      0.926 &   0.917 &     0.921 \\
       & \textbf{ICA} &      0.901 &   0.858 &     0.879 \\
       & \textbf{AE} &      0.934 &   0.916 &     0.922 \\
       & \textbf{DAE} &      0.935 &   0.918 &     0.919 \\
\bottomrule
\end{tabular}
\end{table}

\subsection{Classification on compressed data (Forest Dataset)}

\begin{figure*}
  \centerline{
    \includegraphics[width=\linewidth]{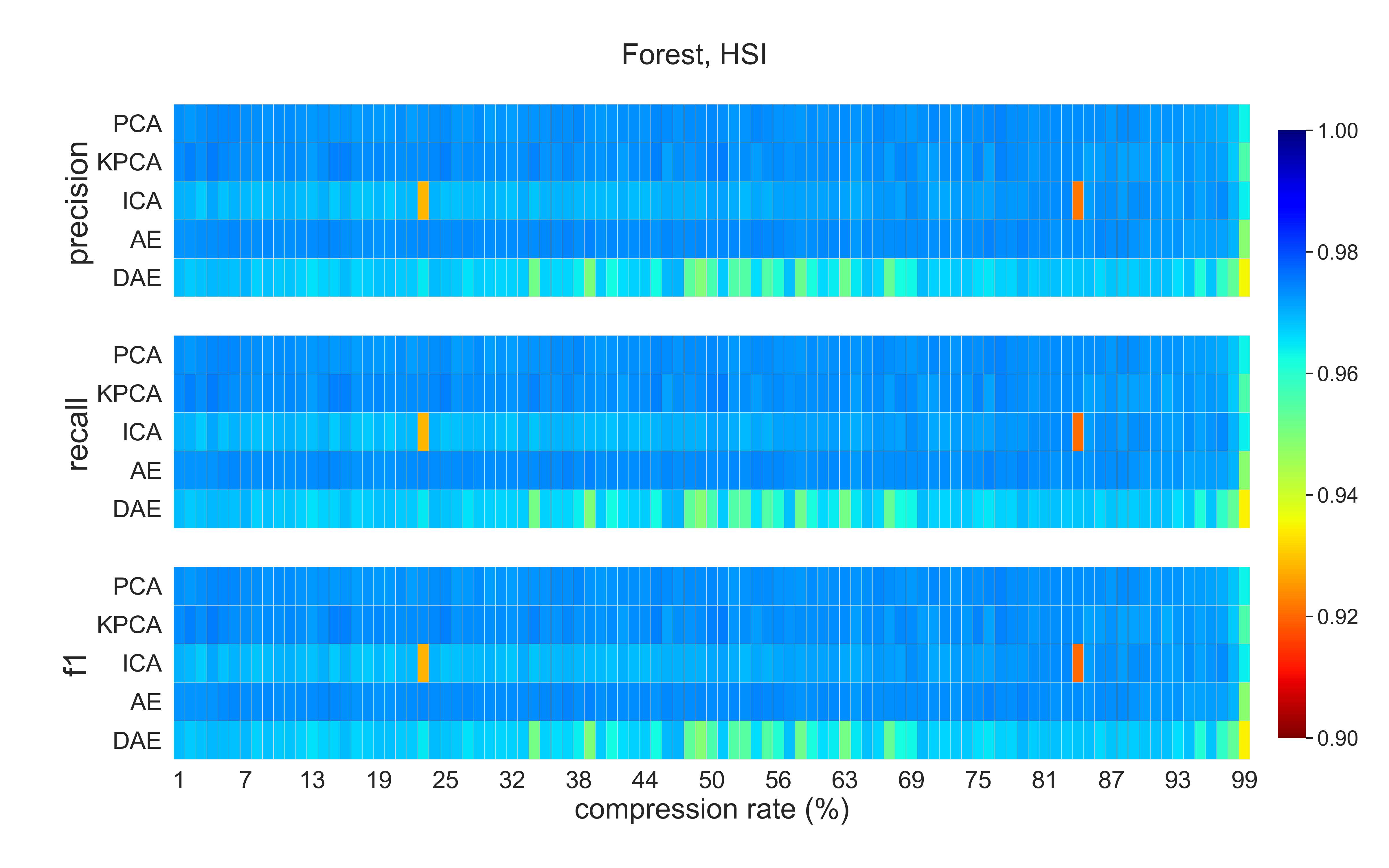}
  }
  \caption{Forest dataset classification scores for all methods for compression rates between 1\% to 99\%.}
  \label{fig:heatmap_Forest_HSI}
\end{figure*}

Figure~\ref{fig:heatmap_Forest_HSI} visualizes classification
\emph{f1-score}, recall, and precision values for compression rates
between 1\% and 99\% on Forest dataset.  DAE and ICA perform poorly on
this dataset.  PCA, KPCA, and AE compression methods achieve good
classification performance on this dataset, where AE outperforming PCA
and KPCA methods at 97\% compression.  Classification performance for
data compressed using ICA method curiously improves with compression
rate.  This merits further investigation.
Table~\ref{table:top_classification_scores_Forest_HSI} shows
\emph{f1-score}, recall, and precision for the Forest dataset at the
95\% compression.  It also includes these values for the RGB baseline.
AE and DAE methods outperform other methods at this level of
compression.

\begin{table}[H]
\centering \footnotesize
\caption{Top classification scores Forest, HSI, compression rate=95\%}
\label{table:top_classification_scores_Forest_HSI}
\begin{tabular}{llrrr}
\toprule
     &     &  precision &  recall &  f1-score \\
\textbf{label} & \textbf{compression} &            &         &           \\
\midrule
\multirow{6}{*}{\textbf{Shadow}} & \textbf{RGB} &      0.971 &   0.968 &     0.970 \\
     & \textbf{PCA} &      0.972 &   0.978 &     0.975 \\
     & \textbf{KPCA} &      0.973 &   0.977 &     0.975 \\
     & \textbf{ICA} &      0.970 &   0.979 &     0.975 \\
     & \textbf{AE} &      0.972 &   0.979 &     0.975 \\
     & \textbf{DAE} &      0.989 &   1.000 &     0.965 \\
\cline{1-5}
\multirow{6}{*}{\textbf{Tree}} & \textbf{RGB} &      0.960 &   0.963 &     0.962 \\
     & \textbf{PCA} &      0.973 &   0.964 &     0.969 \\
     & \textbf{KPCA} &      0.971 &   0.966 &     0.968 \\
     & \textbf{ICA} &      0.974 &   0.962 &     0.968 \\
     & \textbf{AE} &      0.973 &   0.964 &     0.968 \\
     & \textbf{DAE} &      0.964 &   0.992 &     0.956 \\
\bottomrule
\end{tabular}
\end{table}

\subsection{Computational considerations}

\begin{figure*}
  \centerline{
    \includegraphics[width=\linewidth]{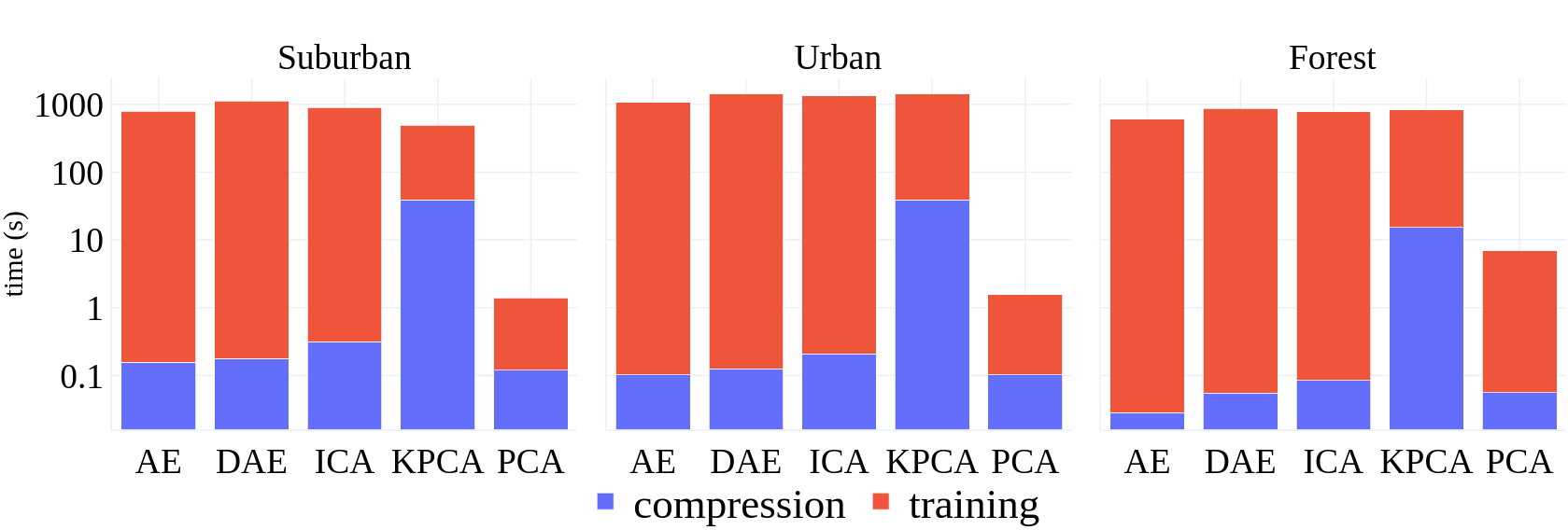}
  }
  \caption{Execution times (seconds) for training and
    compression tasks and all compression algorithms.
  }
  \label{fig:execution_times}
\end{figure*}

The compression methods used in this work need to be trained before
these can be used to compress the incoming 301-dimensional spectral
signal.  The methods only need to be trained once; therefore, the training
time is not important as the compression time (the time it takes these
methods to encode the incoming signal into a low-dimensional space).
Figure~\ref{fig:execution_times} shows compression and training times
for each method.  Notice that KPCA has the largest compression time,
which suggests that KPCA is not well-suited to applications where
smaller runtimes are desireable.

\begin{figure*}
  \centerline{
    \includegraphics[width=\linewidth]{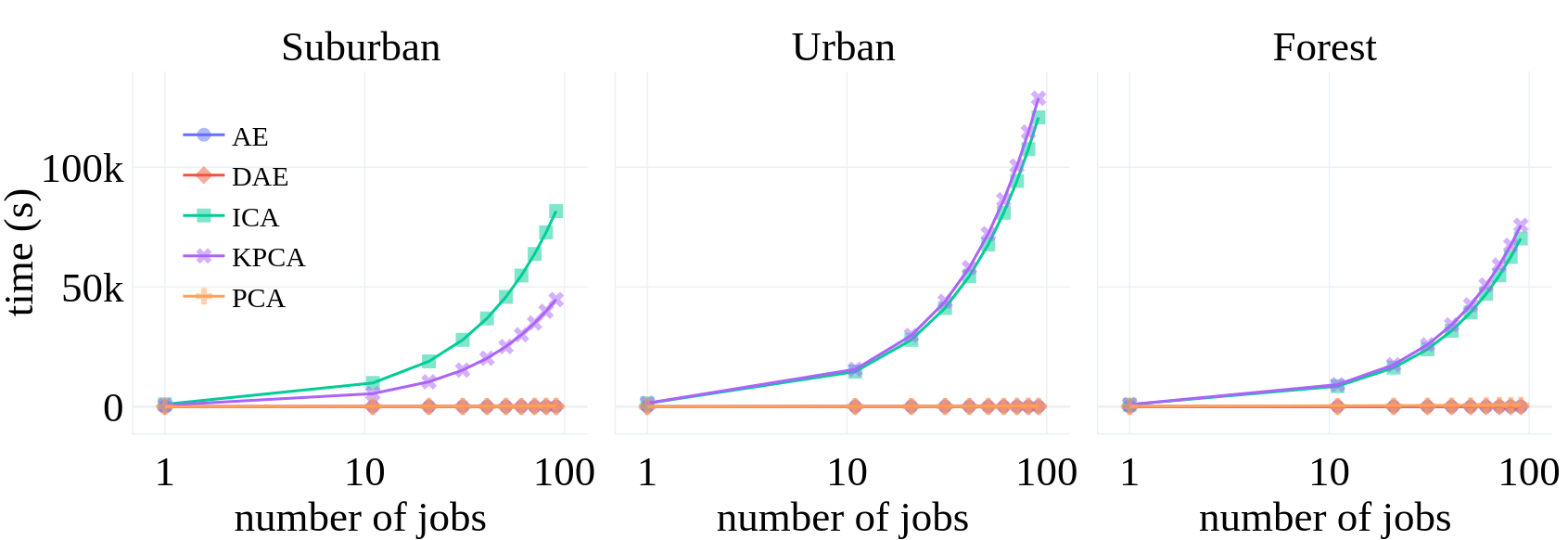}
  }
  \caption{Execution times (in seconds) times the number of
    classification jobs.
  }
  \label{fig:scalability_execution_times}
\end{figure*}

While the training datasets used in this paper fit in memory, one can
imagine a situation where the size of data excludes this
possibility. It is not easy to use PCA, KPCA, and ICA in situations
where training data do not fit in memory.  Deep learning methods,
such as AE and DAE, can be trained in batches; therefore, these
methods can be trained in situations where the entire training data
does not fit in memory.  Figure~\ref{fig:scalability_execution_times}
shows that AE and DAE have better scalability properties than other
compression methods.  This suggests that AE and DAE methods may
be more suited for in-situ applications where computational resources
are usually limited.

\begin{figure*}
  \centerline{
    % trim=left botm right top.
    \includegraphics[width=0.2\linewidth, trim=3cm 0cm 3cm 0cm, clip=true]{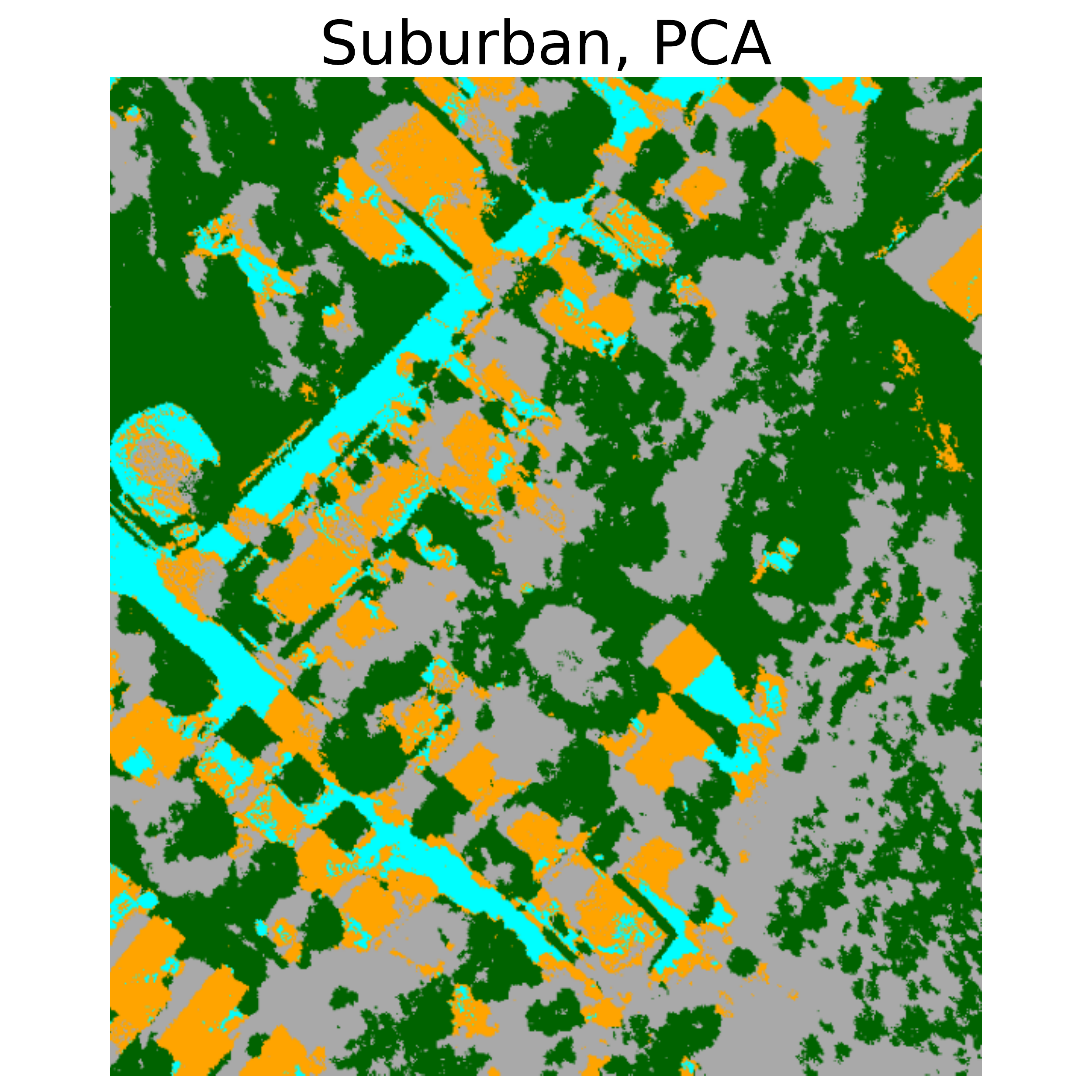}
    \includegraphics[width=0.2\linewidth, trim=3cm 0cm 3cm 0cm, clip=true]{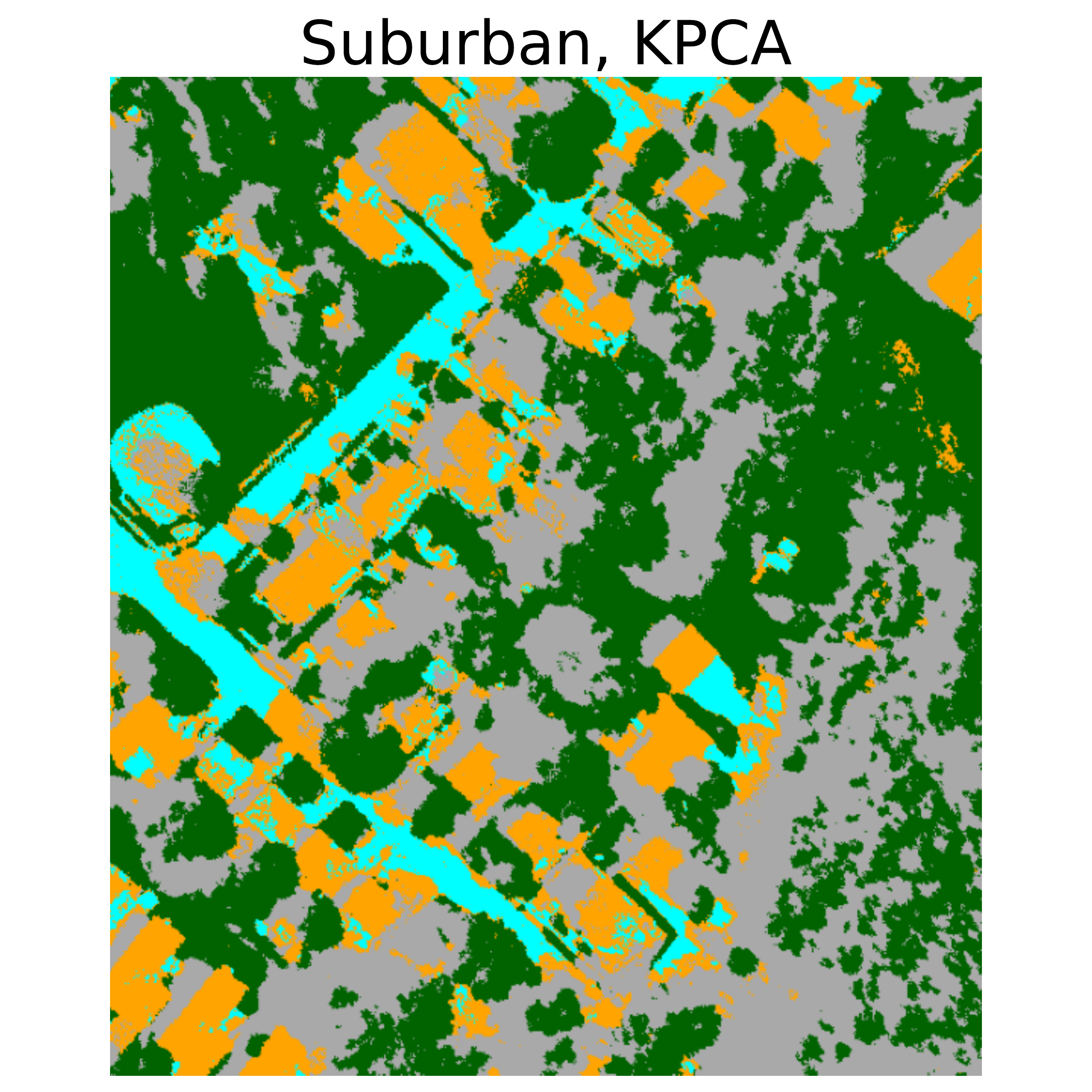}
    \includegraphics[width=0.2\linewidth, trim=3cm 0cm 3cm 0cm, clip=true]{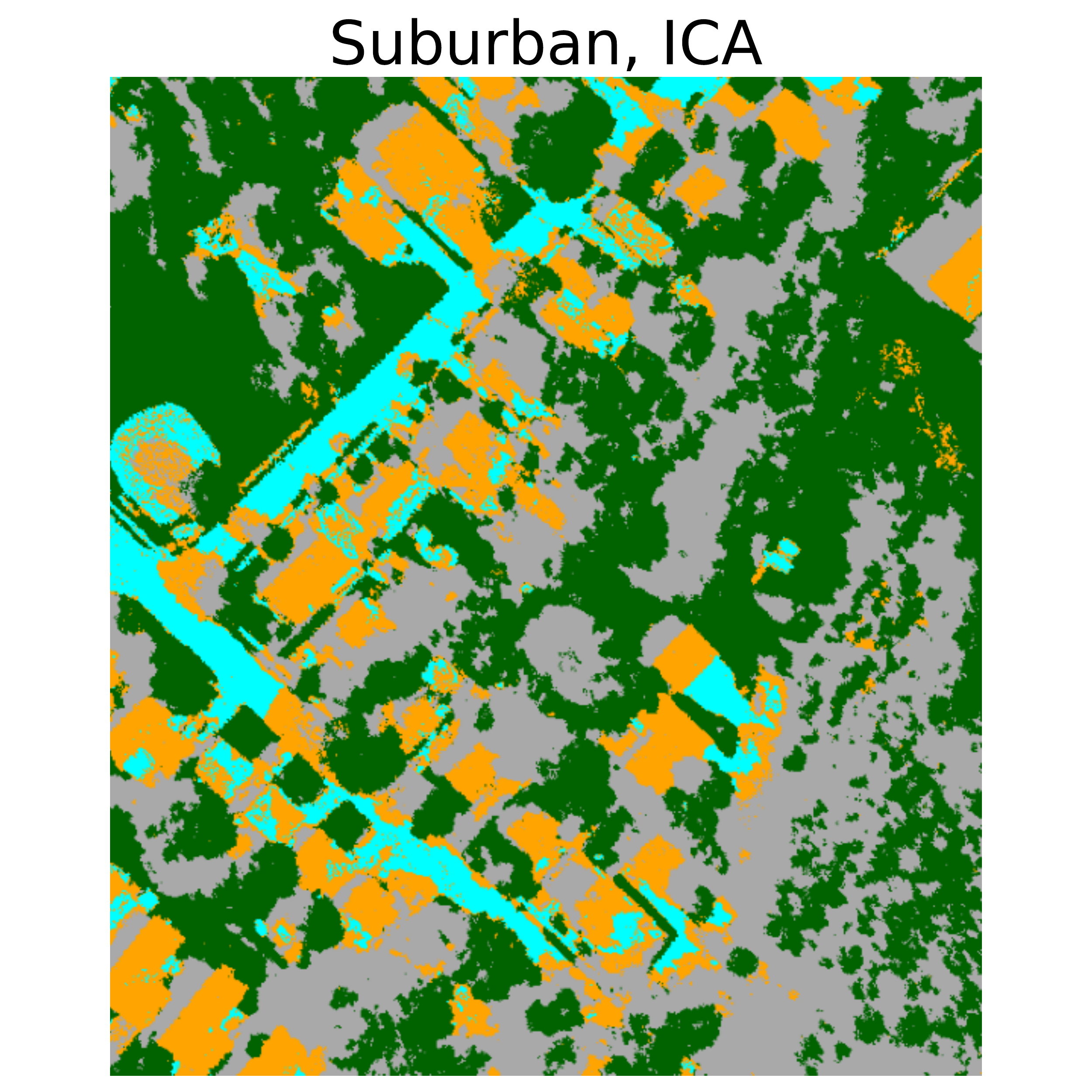}
    \includegraphics[width=0.2\linewidth, trim=3cm 0cm 3cm 0cm, clip=true]{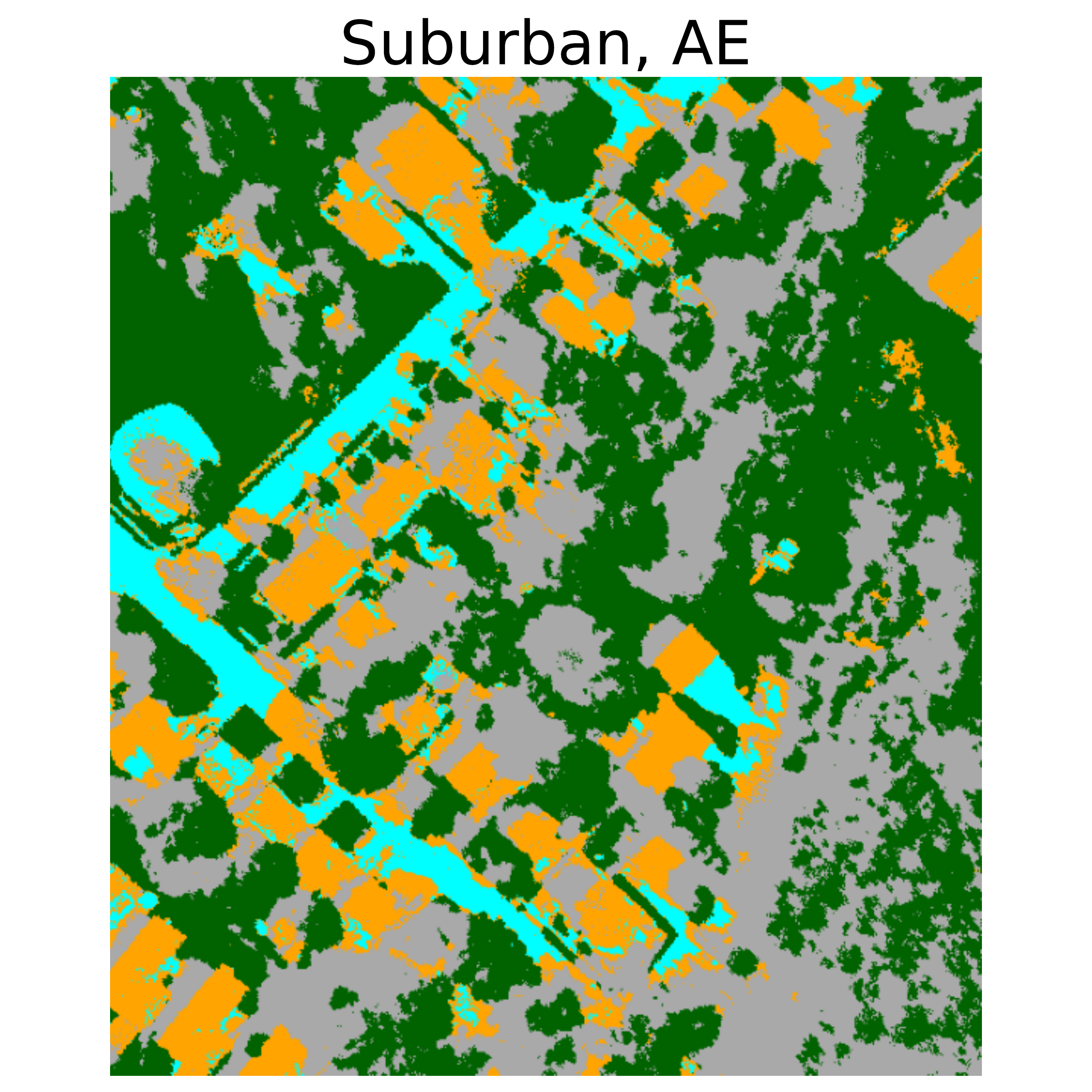}
    \includegraphics[width=0.2\linewidth, trim=3cm 0cm 3cm 0cm, clip=true]{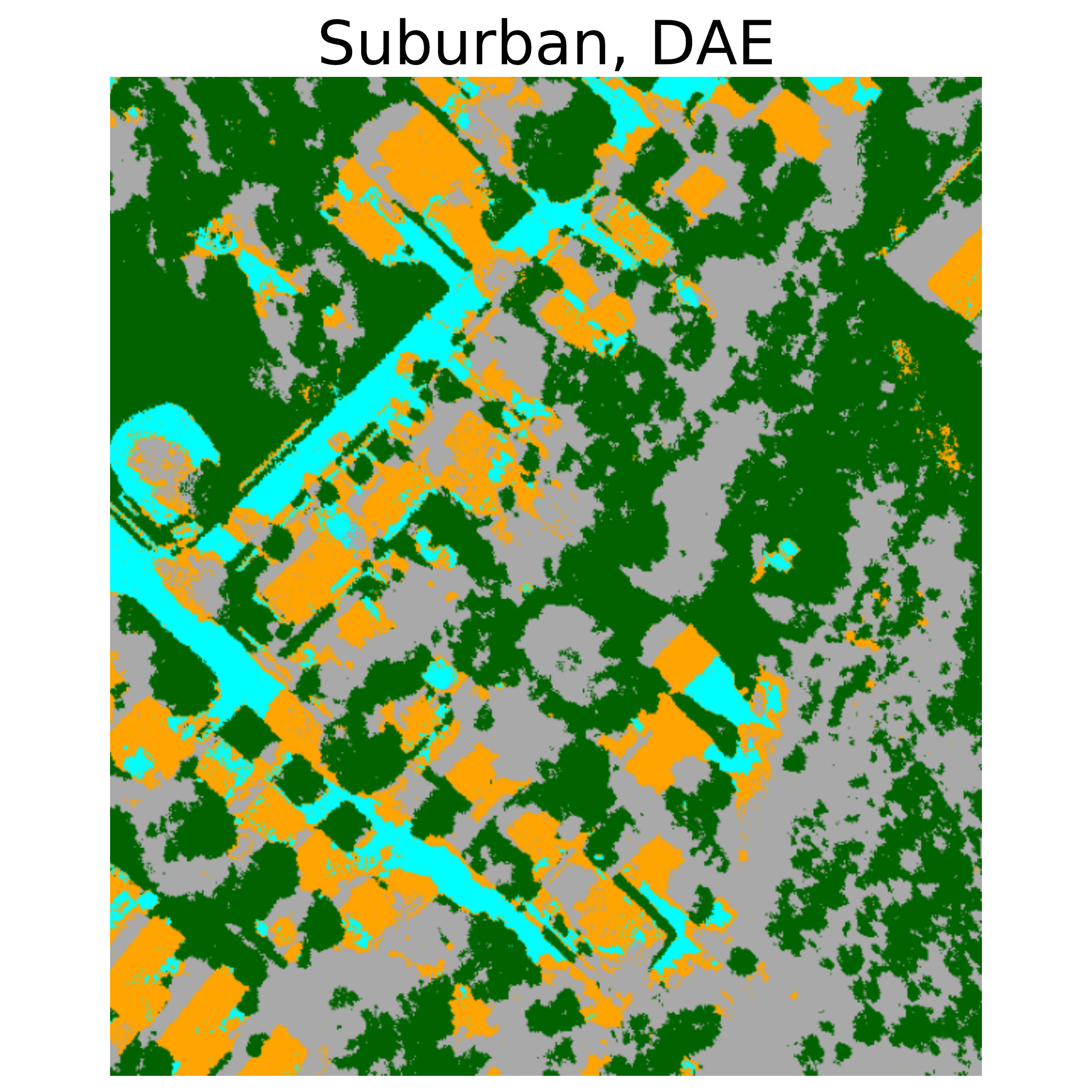}
  }
  \centerline{
    \includegraphics[width=0.2\linewidth, trim=3cm 0cm 3cm 0cm, clip=true]{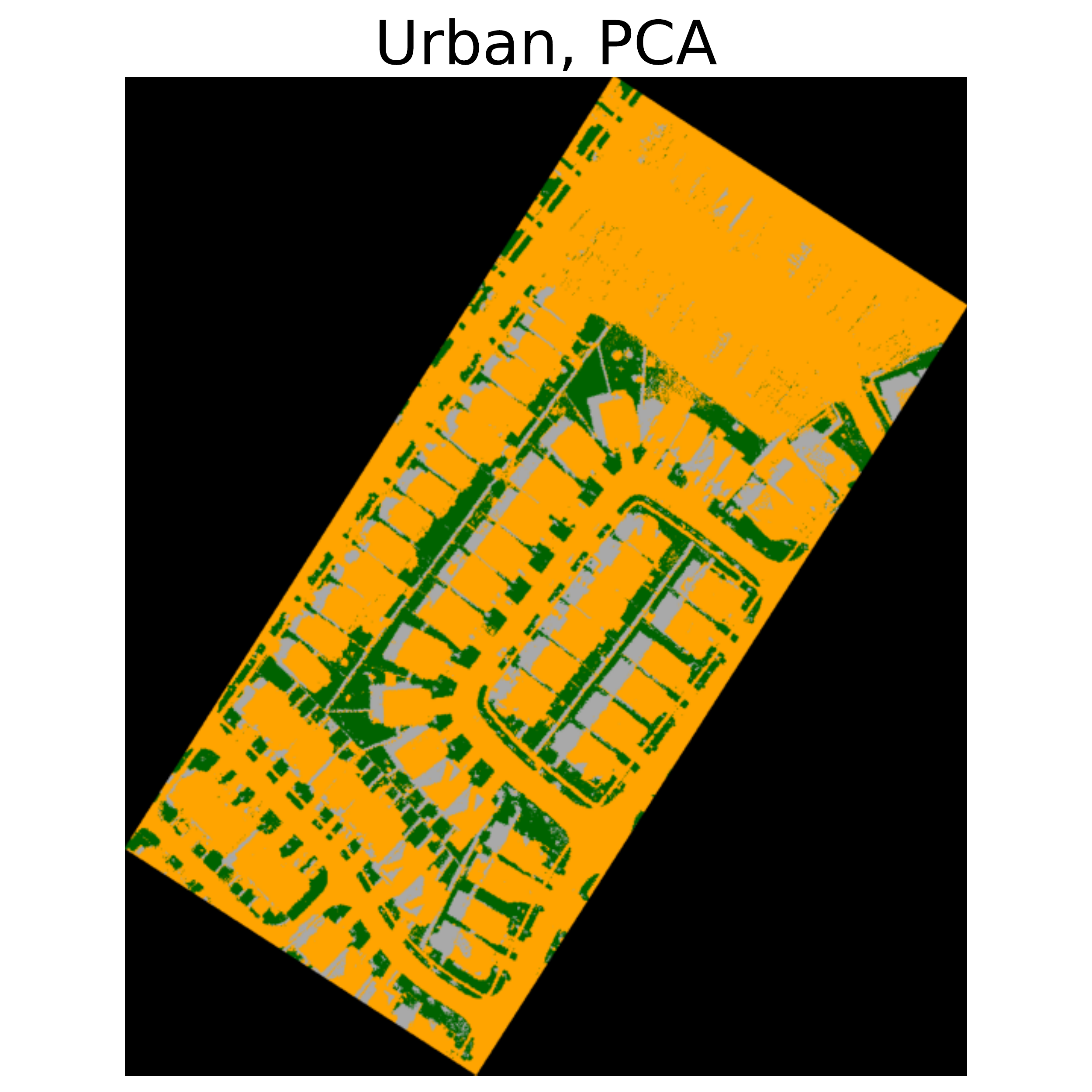}
    \includegraphics[width=0.2\linewidth, trim=3cm 0cm 3cm 0cm, clip=true]{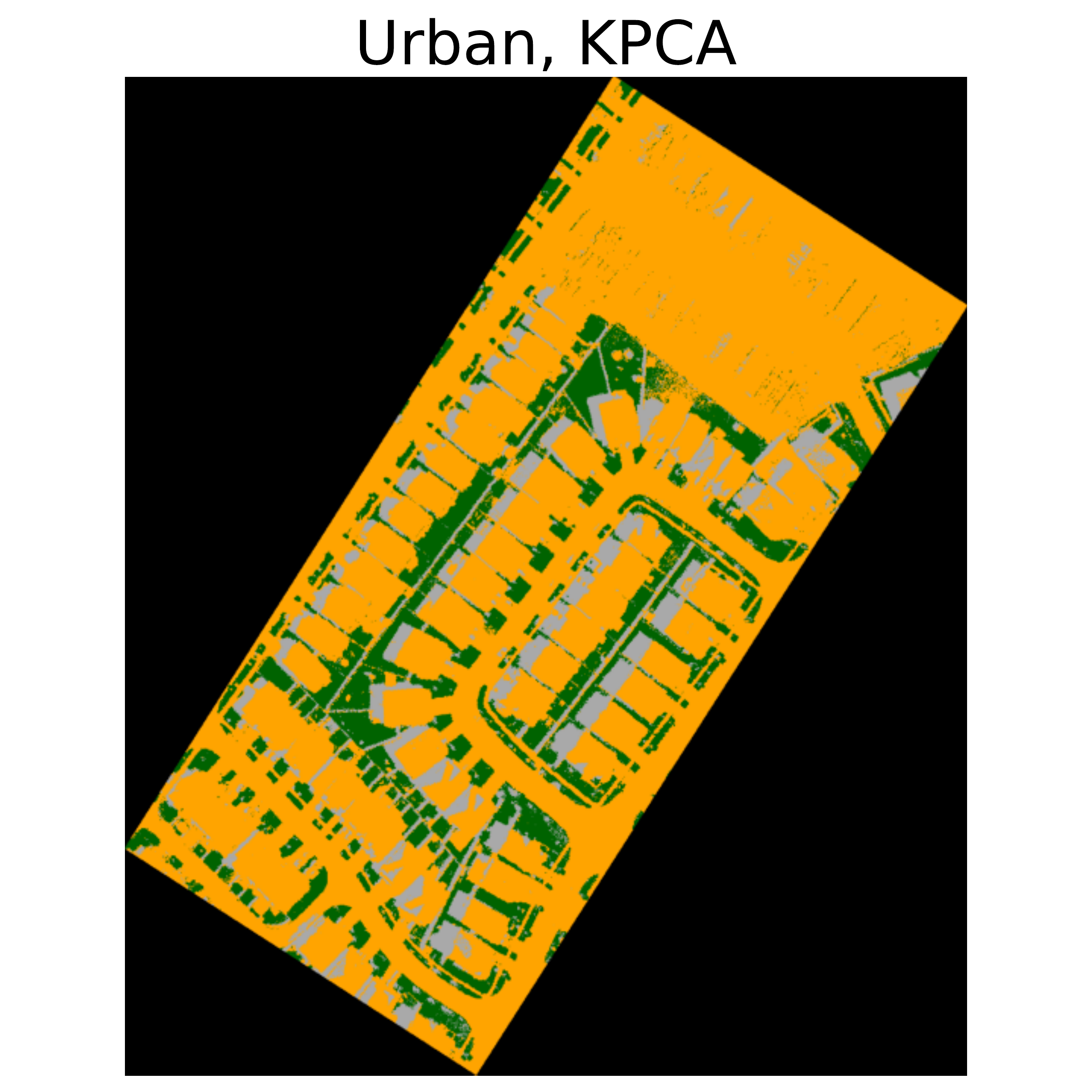}
    \includegraphics[width=0.2\linewidth, trim=3cm 0cm 3cm 0cm, clip=true]{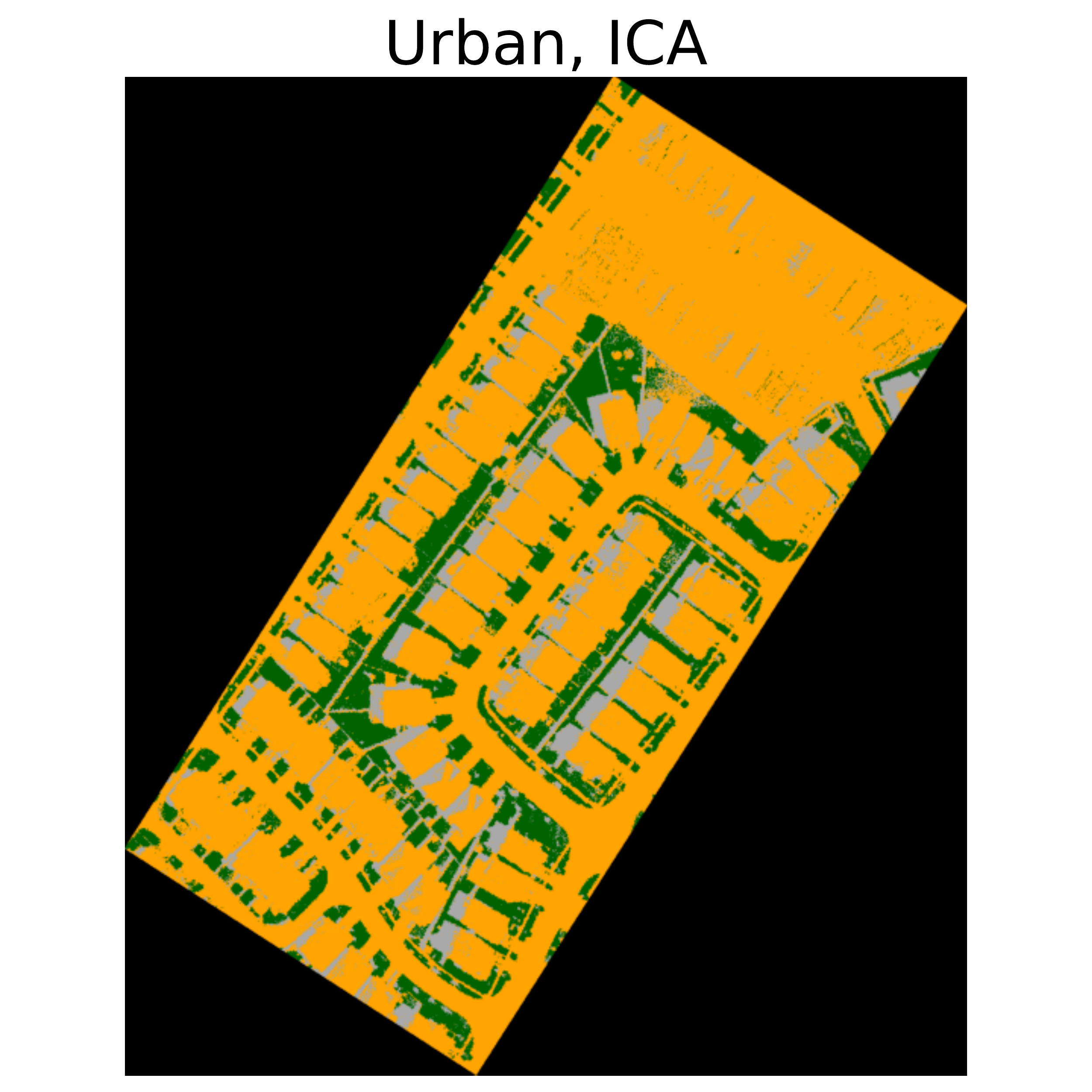}
    \includegraphics[width=0.2\linewidth, trim=3cm 0cm 3cm 0cm, clip=true]{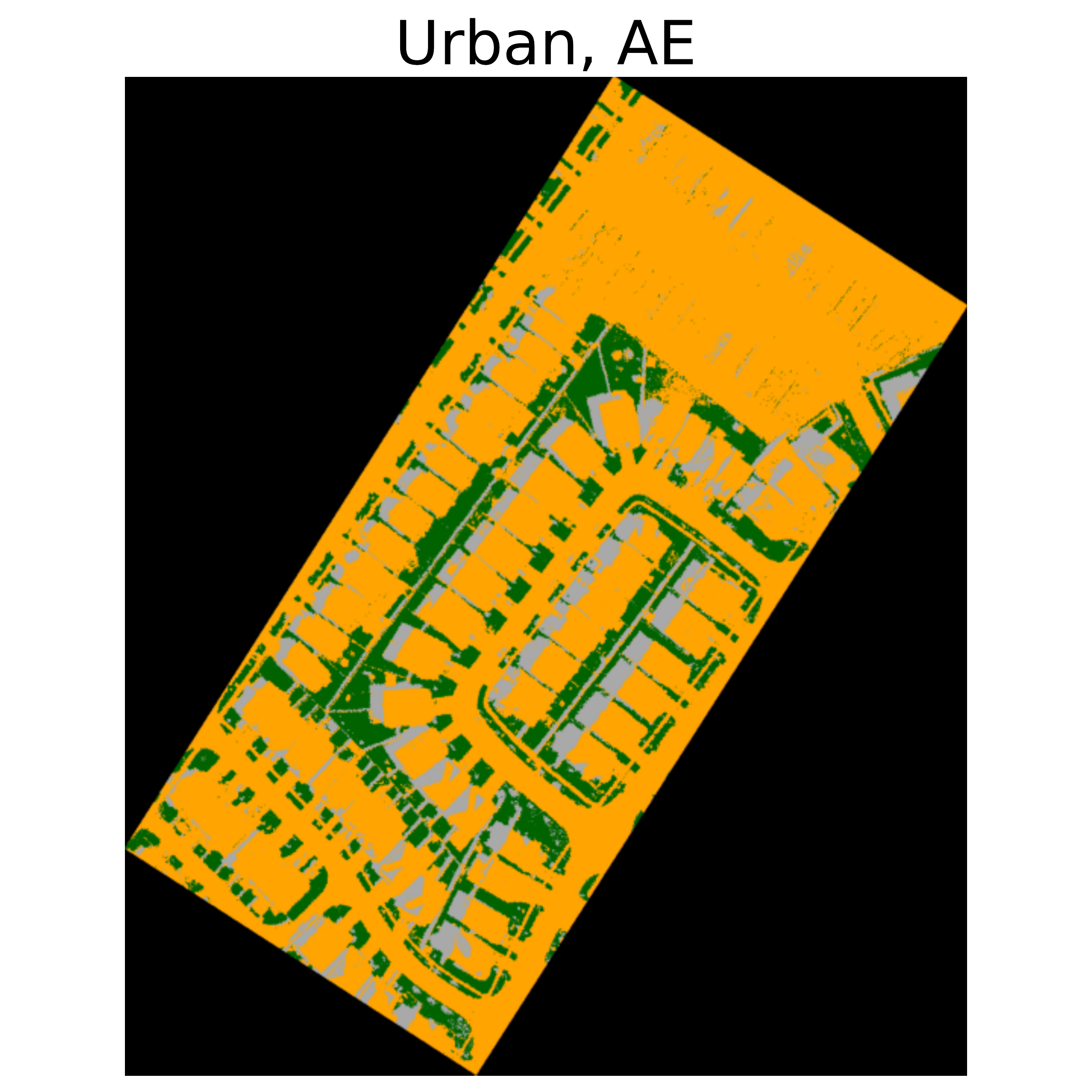}
    \includegraphics[width=0.2\linewidth, trim=3cm 0cm 3cm 0cm, clip=true]{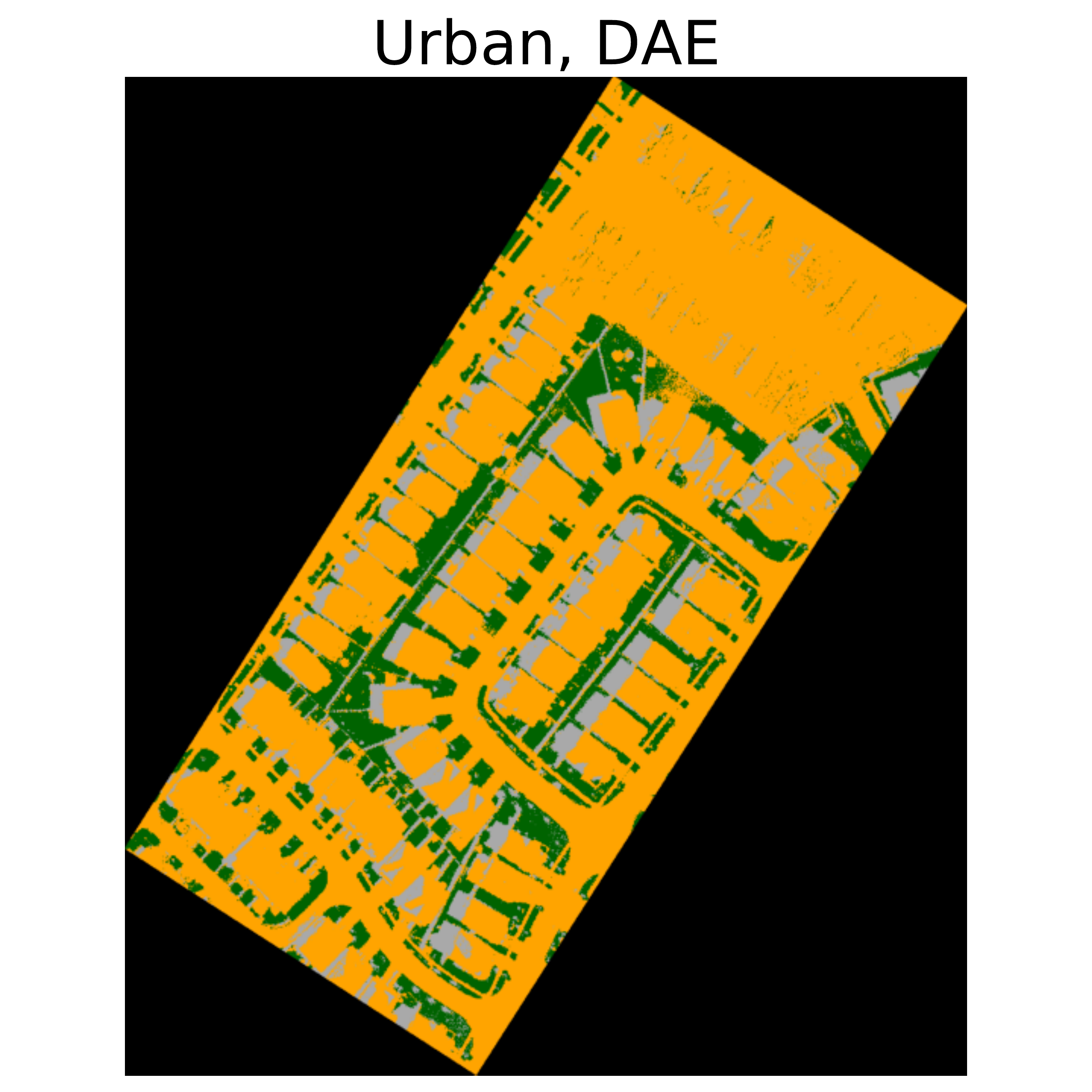}
  }
  \centerline{
    \includegraphics[width=0.2\linewidth, trim=2cm 0cm 2cm 0cm, clip=true]{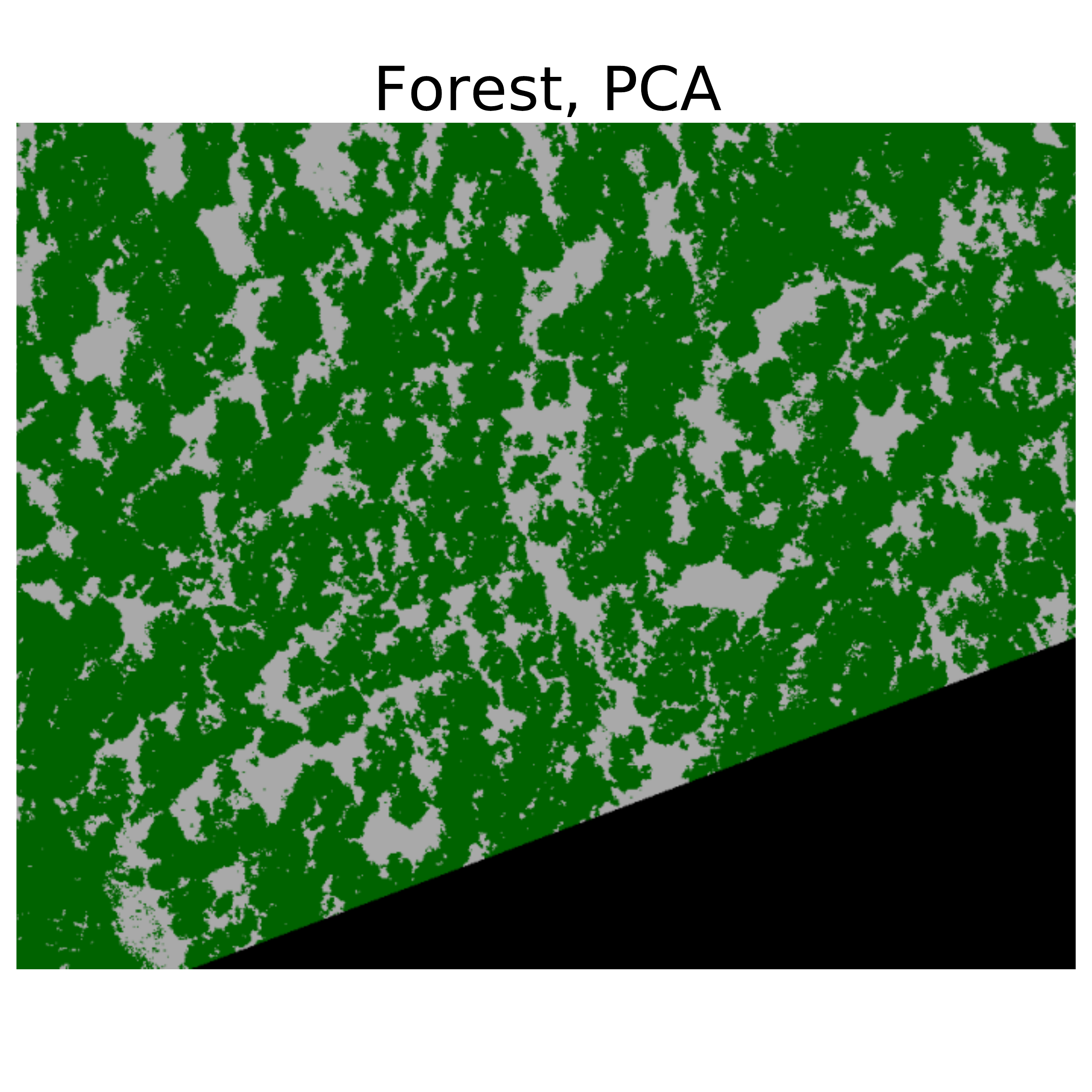}
    \includegraphics[width=0.2\linewidth, trim=2cm 0cm 2cm 0cm, clip=true]{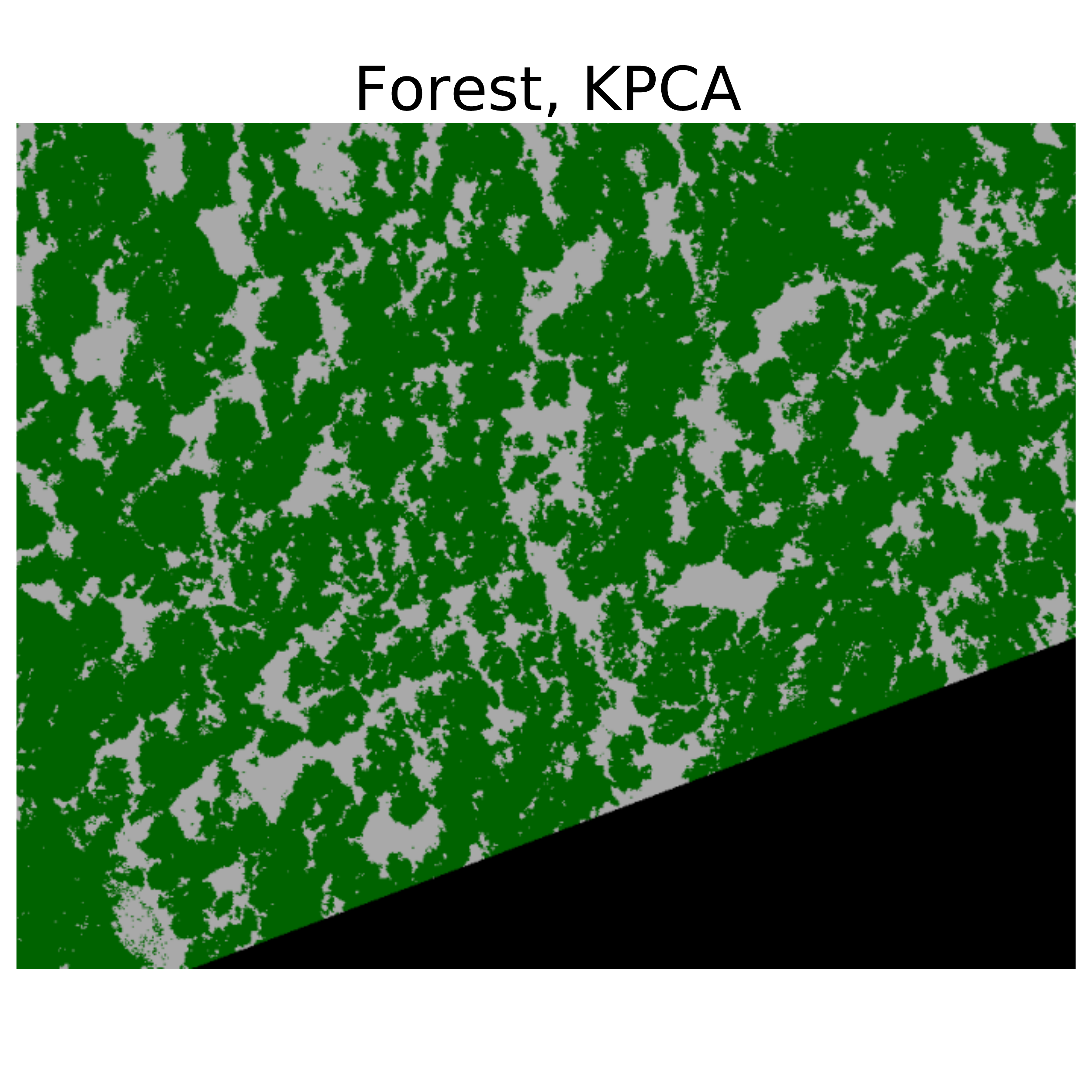}
    \includegraphics[width=0.2\linewidth, trim=2cm 0cm 2cm 0cm, clip=true]{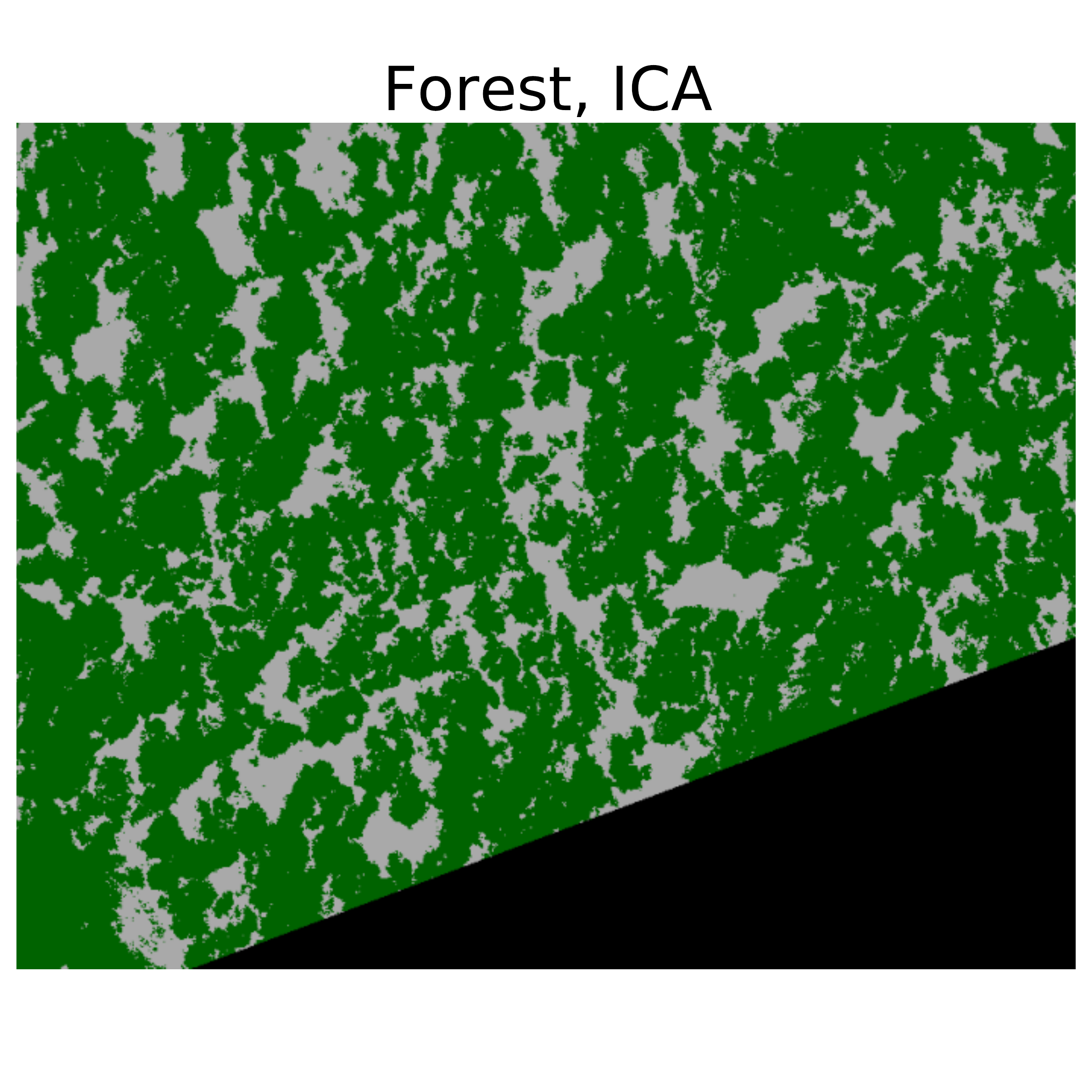}
    \includegraphics[width=0.2\linewidth, trim=2cm 0cm 2cm 0cm, clip=true]{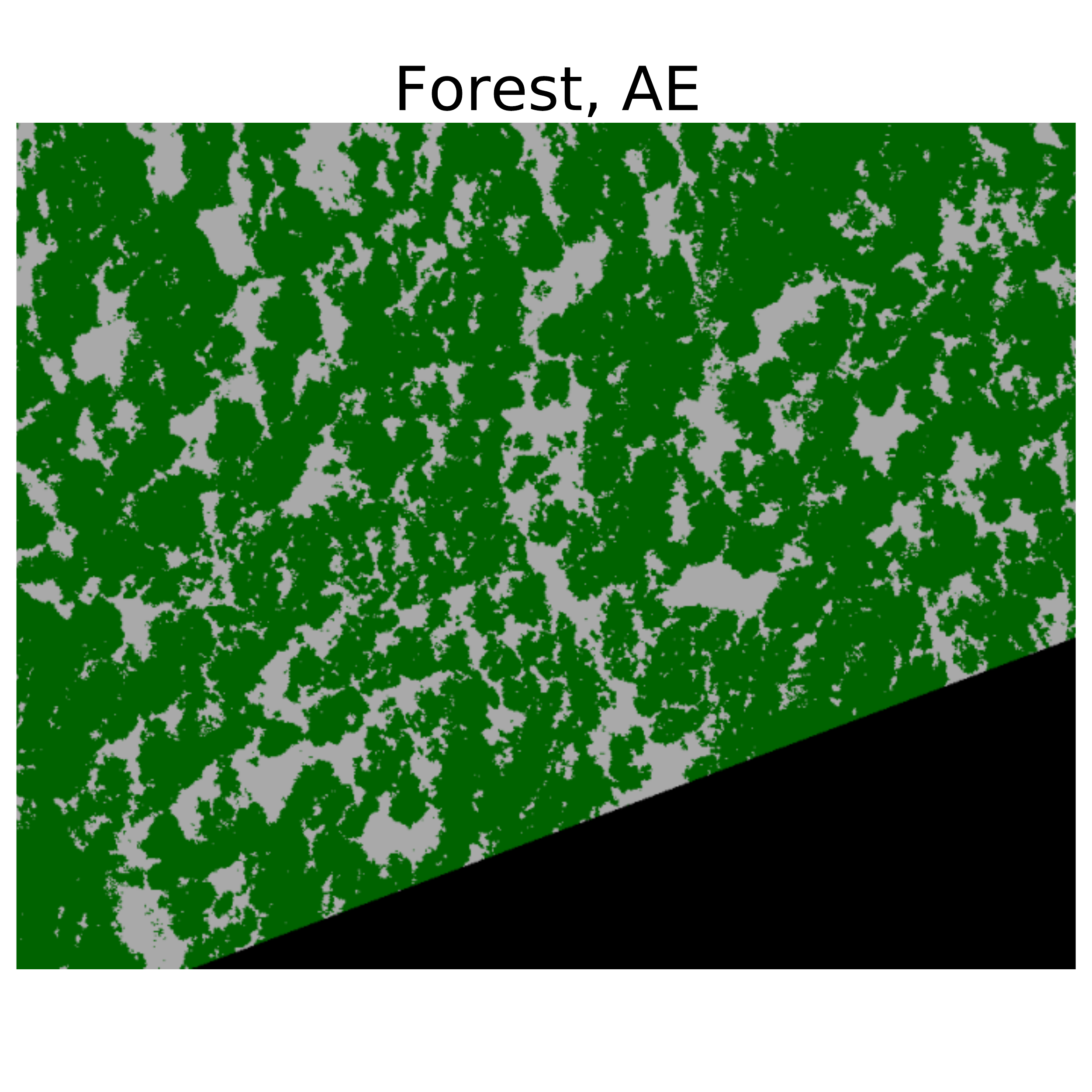}
    \includegraphics[width=0.2\linewidth, trim=2cm 0cm 2cm 0cm, clip=true]{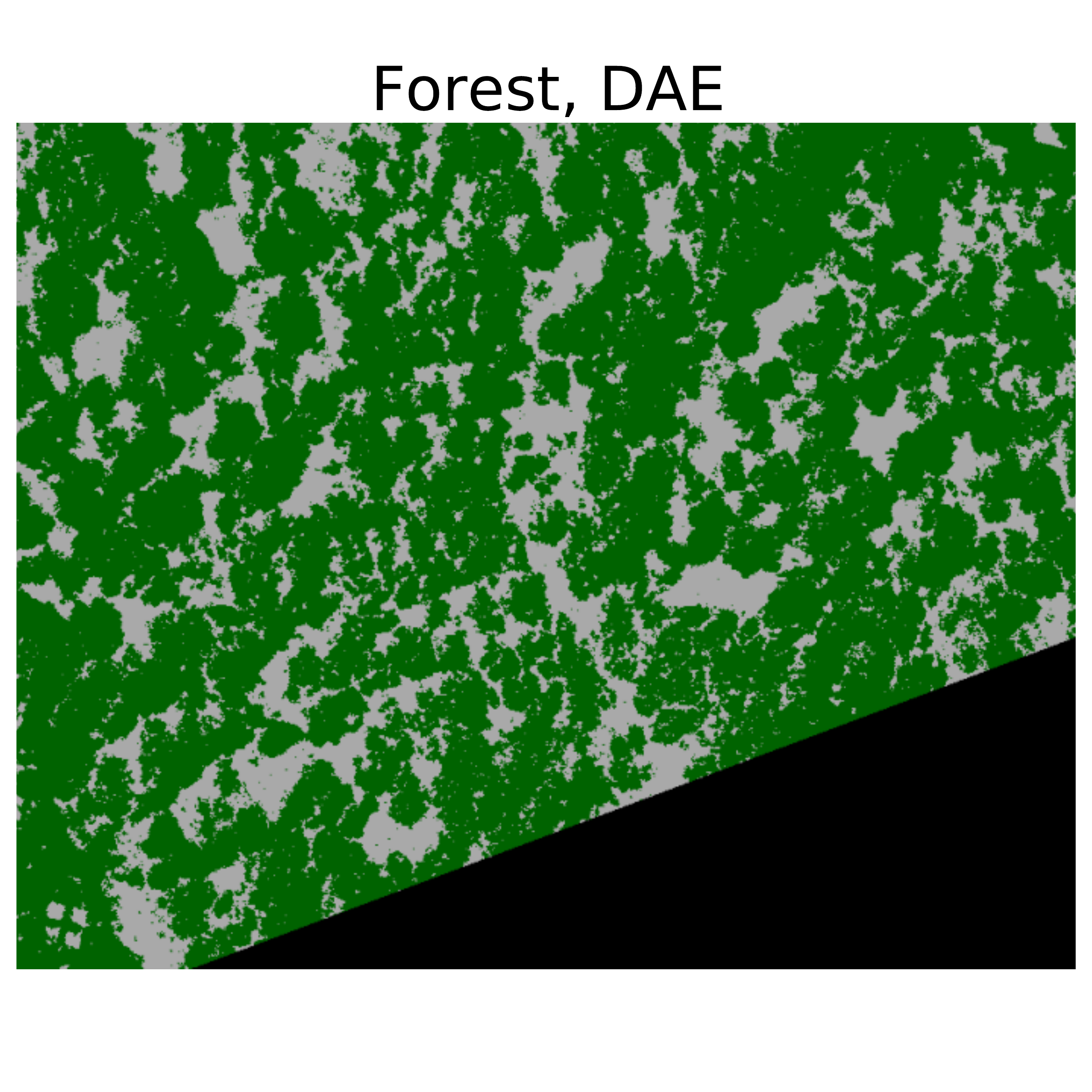}
  }
  \caption{Classification images with spectra compressed to 95\% of original size.}
  \label{fig:classification-images}
\end{figure*}

\section{Conclusion}
\label{sec:conclusion}

Hyperspectral pixels contain two orders of magnitude more information
than ordinary RGB pixels, and it is often possible to carry out
analysis tasks, such as segmentation and classification, without using
the complete spectral signal.  As a result, dimensionality reduction
techniques, such as PCA, KPCA, ICA, AE, and DAE, are widely employed
as a first step in the overall hyperspectral image analysis pipeline.
This paper presents a systematic study that investigates the effects
of compression on hyperspectral pixel classification.  Specifically,
we implemented five compression methods---PCA, KPCA, ICA, AE, and DAE---and
used these to compress 301-band hyperspectral pixels from three different
hyperspectral image datasets. The compression rates varied from
99\% to 1\%.  Gradient Boosted Decision Tree (XGBoost) classifiers were
trained for each (compression method, rate, and dataset) yielding a total
of 1470 classifiers. Reconstructions scores, classification accuracy,
and runtimes for each (compression method, compression rate, classifier, dataset)
were recorded to perform an empirical study on the effects of compression
on hyperspectral pixel-level classification.

We found that PCA, KPCA, and ICA post lower signal reconstruction
errors; however, these methods achieve lower classification scores
when the compression rate is greater than 95\%. AE and DAE methods post better
classification accuracy at compression rates higher than 98\%. Noise reduction
filtering, which is a common signal preprocessing step for hyperspectral images,
is not needed when using DAE for compression.  We also captured the runtime
performance of different compression methods, and we found that AE and DAE  methods
are well-suited for resource-constrained, in-situ settings.  Our results suggest
that the choice of a compression method and compression rate is an important
consideration when designing a hyperspectral pixel classification pipeline.

In the analysis presented in this paper, each hyperspectral pixel is treated independently.
In the future, we plan to study Markov Random Field approaches to capture
the relationship between neighbouring pixels during hyperspectral pixel classification
and image segmentation tasks.

\section*{Acknowledgments}

We acknowledge the support of the Natural Sciences and Engineering Research Council
of Canada (NSERC) through the NSERC Discovery Program for the funding of the hyperspectral
image acquisition mission and image preprocessing facility (RGPIN-386183 awarded to Dr. Yuhong He),
and for the Visual Computing Lab of the Ontario Tech University (RGPIN-2020-05159,
awarded to Dr. Faisal Z. Qureshi).

% \bibliography{../refs}

\end{document}